\documentclass{article}
\usepackage{graphicx}
\usepackage{subcaption}
\usepackage{booktabs}
\usepackage{microtype}
\usepackage{algorithmic}
\usepackage{amsfonts}
\usepackage{amsmath}
\usepackage{amssymb}
\usepackage{mathtools}
\usepackage{amsthm}
\usepackage{multirow}
\usepackage{makecell}
\usepackage{orcidlink}
\usepackage{array}
\usepackage{tabularx}
\usepackage{pifont}
\usepackage{bm}
\usepackage{balance}
\usepackage{ulem}[normal]
\usepackage[capitalize,noabbrev]{cleveref}
\theoremstyle{plain}
\theoremstyle{remark}
\usepackage[textsize=tiny]{todonotes}

\usepackage{hyperref}
\usepackage[preprint]{icml2026}
\usepackage{orcidlink}

\renewcommand{\algorithmicrequire}{ \textbf{Input:}} 
\renewcommand{\algorithmicensure}{ \textbf{Output:}} 

\icmltitlerunning{SDM: A Powerful Tool for Evaluating Model Robustness}

\begin{document}

\twocolumn[
  \icmltitle{SDM: A Powerful Tool for Evaluating Model Robustness}
  \icmlsetsymbol{equal}{*}
  \begin{icmlauthorlist}
    \icmlauthor{Xinlei Liu \orcidlink{0000-0001-6586-4457}}{equal,x,y,z}
    \icmlauthor{Tao Hu \orcidlink{0000-0001-7641-5622}}{equal,x,y,z}
    \icmlauthor{Jichao Xie}{x}
    \icmlauthor{Peng Yi}{u}
    \icmlauthor{Hailong Ma}{x,y,z}
    \icmlauthor{Baolin Li}{x,y,z}
  \end{icmlauthorlist}

  \icmlaffiliation{x}{Information Engineering University, Zhengzhou, China}
  \icmlaffiliation{y}{Key Laboratory of Cyberspace Endogenous Safety \& Security of Henan Province, Zhengzhou, China}
  \icmlaffiliation{z}{Key Laboratory of Cyberspace Security Ministry of Education of China, Zhengzhou, China}
  \icmlaffiliation{u}{Songshan Laboratory, Zhengzhou, China}

  \icmlcorrespondingauthor{Tao Hu}{hutaondsc@163.com}

  \icmlkeywords{Adversarial Robustness, Adversarial Attack, Adversarial Example, AI Security, Image Classification}

  \vskip 0.3in
]

\printAffiliationsAndNotice{\icmlEqualContribution}

\begin{abstract}
Gradient-based attacks are important methods for evaluating model robustness. However, since the proposal of APGD, it has been difficult for such methods to achieve significant breakthroughs. To achieve such an effect, we first analyze the issue of ``high-loss non-adversarial examples" that degrades attack performance in previous methods, and prove that this issue arises from inappropriate objectives for adversarial example generation. Subsequently, we reconstruct the objective as ``maximizing the difference between the non-ground-truth label probability upper bound and the ground-truth label probability", and proposes a novel and powerful gradient-based attack method named \textbf{Sequential Difference Maximization} (\textbf{SDM}). SDM establishes a three-layer optimization framework of ``cycle-stage-step". It adopts the negative probability loss function and the \textbf{Directional Probability Difference Ratio} (\textbf{DPDR}) loss function in the initial and subsequent optimization stages, respectively, and approaches the ideal objective of adversarial example generation via stage-wise sequential optimization. Experiments demonstrate that compared with previous state-of-the-art methods, SDM not only achieves stronger attack performance but also exhibits superior cost-effectiveness. The code is available at \url{https://github.com/X-L-Liu/ICML-SDM}.
\end{abstract}

\section{Introduction}
\label{Sec1}

Adversarial examples are constructed by adding imperceptible adversarial perturbations to natural examples, which can cause well-trained deep neural networks (DNNs) to make incorrect judgments \cite{Goo15,Akh21}. Taking image classification tasks as an example, adversarial example attacks (adversarial attacks) can induce target models to output misclassification results with high confidence, posing severe threats to downstream tasks such as face recognition and object detection \cite{Liu24,Bar24,PAN}. However, from another perspective, adversarial attacks also profoundly reveal the limitations of DNNs in terms of robustness and essential understanding \cite{Cro20}. This prompts researchers to explore ways to enable models to learn more intrinsic and stable feature representations, rather than merely capturing the statistical patterns on the surface of data.

An important application scenario of adversarial attacks is the white-box setting. In this scenario, attackers have access to all information of the defender, including model parameters and defense strategies, which represents the most stringent evaluation condition. Currently commonly used white-box attacks mainly include gradient-based methods \cite{Goo15}, optimization-based methods \cite{Car17}, and decision boundary analysis-based methods \cite{Sey16}. Compared with other approaches, gradient-based methods only require computing the gradient of the loss with respect to the input, featuring clearer conceptual interpretability and simpler code implementation. Among them, the representative Projected Gradient Descent (PGD) \cite{Mad18} and its advanced variants (e.g., Auto-PGD, APGD) \cite{Cro20} have almost become the universal benchmark for evaluating model robustness, owing to their standardized perturbation projection constraints, moderate computational cost and strong attack performance \cite{Bad24}.

However, we identify a issue of ``high-loss non-adversarial examples" arising from the unreasonable optimization objective employed by these methods: for certain natural examples, the high-loss examples generated by PGD fail to achieve successful attacks, whereas low-loss examples generated by alternative methods can achieve successful attacks instead. This problem compromises the attack effectiveness of methods such as PGD, thereby undermining their validity in assessing model robustness. 

To address the aforementioned problem, we first reconstruct the adversarial example generation objective as ``maximizing the difference between the non-ground-truth label probability upper bound and the ground-truth label probability". Then, We propose Sequential Difference Maximization (SDM) to optimize this objective. SDM decomposes the solution process into multiple stages: (1) The initial stage aims to minimize the ground-truth label probability, thereby compressing the solution space for reduced complexity. (2) Subsequent stages introduce the Directional Probability Difference Ratio (DPDR) loss function to increase the non-ground-truth label probability upper bound. Each stage takes the previous stage's optimal solution as the initial solution, and all stages are optimized strictly in the set order. Experimental results show that SDM outperforms previous methods in attack performance and cost-effectiveness. We summarize the main contributions as follows:

\begin{itemize}
\item We propose a gradient-based adversarial attack method known as \textbf{SDM}, which efficiently generates adversarial examples by reconstructing the optimization objective and introducing \textbf{DPDR} loss function in multiple ordered optimization stages.
\item We identify the issue of ``\textbf{high-loss non-adversarial examples}", which reveals the irrationality in the setting of optimization objectives and loss functions in some prior adversarial attack methods.
\item Experimental results demonstrate that SDM achieves superior attack performance, cost-effectiveness, and anti-interference capability, outperforming the prior state-of-the-art methods significantly.
\end{itemize}

\section{Related Work}
\label{Sec2}

\subsection{Adversarial Example}

Let \(f\) be a multiclass classifier mapping inputs from \(\mathbb{R}^d\) to output scores in \(\mathbb{R}^K\) (\(K>2\)). \(d\) and \(K\) represent the dimension of the input space and the total number of labels, respectively. For a natural example \(\boldsymbol{x}\in[0, 1]^d \subset\mathbb{R}^d\) with the ground-truth label \(y\), the model produces a vector \(\left\{f_{1}(\boldsymbol{x}), f_{2}(\boldsymbol{x}), \dots, f_{K}(\boldsymbol{x})\right\}\) as the output score. 

Adversarial examples refer to input examples that can actively mislead a target classification model into making incorrect predictions. If no specific direction of the incorrect prediction is designated, they are termed untargeted adversarial examples. As the core research object of this paper, untargeted adversarial examples are directly referred to as ``adversarial examples" in subsequent content. An ideal adversarial example \(\boldsymbol{x}'\) can be formulated as:

\begin{equation}
\operatorname*{\arg\max}_{k\in\{1,\dots,K\}} f_k(\boldsymbol{x}') \neq y \quad \text{s.t.} \quad \left\|\boldsymbol{x}'-\boldsymbol{x}\right\|_p \leq \epsilon,
\label{Equ1}
\end{equation}

\noindent where \(\epsilon\) and \(\left\|\cdot\right\|_p\) represent the adversarial perturbation budget and the \(\bm{\ell}_p\)-norm constraint, respectively.

\subsection{Gradient-Based Adversarial Attack}

Gradient-based methods represent the most commonly adopted approach for adversarial attacks (i.e., adversarial example generation). In the Fast Gradient Sign Method (FGSM) method \cite{Goo15}, \textit{Goodfellow} et al. first formalized the objective of adversarial example generation as follows:

\begin{equation}
\min_{\boldsymbol{x}'} f_y(\boldsymbol{x}') \quad \text{s.t.} \quad \left\|\boldsymbol{x}' - \boldsymbol{x}\right\|_p \leq \epsilon.
\label{Equ2}
\end{equation}

Compared with Equation~\ref{Equ1}, which exhibits high nonlinearity and poses great optimization challenges \cite{Chr14}, Equation~\ref{Equ2} clearly offers a more explicit optimization pathway. Specifically, FGSM first calculates the Cross Entropy (CE) loss between the model’s output logits and the ground-truth label, then amplifies the adversarial perturbation along the direction of the partial derivative (gradient) of this loss with respect to the input example, thereby generating an adversarial example. FGSM is widely regarded as the foundational work for gradient-based adversarial attacks.

In subsequent research, the PGD and Basic Iterative Method (BIM) \cite{Kur17} extended the single-step perturbation to multi-step iteration, generating more potent adversarial examples through a series of small-step attack and projection operations. However, their optimization processes may still converge to local optima and are highly sensitive to hyperparameters. APGD \cite{Cro20} recognizes that the fixed step size in PGD is suboptimal and that the algorithm is insensitive to the adversarial perturbation budget. Thus, APGD eliminates the need for preset hyperparameters such as step size and iteration count; instead, it dynamically and adaptively adjusts the step size, enabling more robust and efficient generation of more threatening adversarial examples. In addition, gradient-based attacks include other representative methods such as the Elastic-Net Attacks to DNNs (EAD) \cite{Che18} and Fast Adaptive Boundary attack (FAB) \cite{CroMin20}, which enhance attack performance by introducing mixed-norm constraints and direct decision boundary optimization.

\section{Motivation}
\label{Sec3}

\begin{table*}[t]
\caption{Scores and probabilities of two adversarial examples generated from a natural example.}
\label{Tab1}
  \centering
  \newcolumntype{C}{>{\centering}X}
  \newcolumntype{L}[1]{>{\centering\arraybackslash}p{#1}}
  \renewcommand{\arraystretch}{0.99}
  \scalebox{0.87}{
  \begin{tabularx}{1.12\linewidth}{C|L{0.85cm}|*{10}{@{\hspace{0.1em}}L{1.12cm}@{\hspace{0.1em}}}|L{1.03cm}|L{1.37cm}|L{1.33cm}}
    \toprule
    \multirow{2}{*}{\makebox[0pt][c]{Input}} & \multirow{2}{*}{\makebox[0pt][c]{Output}} & \multicolumn{10}{c|}{\(k\)} & \multirow{2}{*}{\makebox[0pt][c]{CE Loss}} & \multirow{2}{*}{\makebox[0pt][c]{\scalebox{0.95}[1]{Pred. Label}}} & \multirow{2}{*}{\makebox[0pt][c]{\scalebox{0.95}[1]{Att. Result}}}\\
    & & 1 & 2 & 3 & \textbf{4} (\(y\)) & 5 & 6 & 7 & 8 & 9 & 10 & &\\
    \midrule
    \multirow{2}{*}{\makebox[0pt][c]{\(\boldsymbol{x}'_{(1)}\)}} & \makebox[0pt][c]{\(S_k\)} & 0.314 & -1.267 & -0.126 & \textbf{1.438} & 0.264 & 1.036 & 0.191 & -0.118 & -0.498 & -1.041 & \multirow{2}{*}{1.196} & \multirow{2}{*}{4} & \multirow{2}{*}{\textcolor{red}{Failed}}\\
    & \makebox[0pt][c]{\(P_k\)} & 9.83\% & 2.02\% & 6.33\% & \textbf{30.25\%} & 9.35\% & 20.24\% & 8.69\% & 6.38\% & 4.36\% & 2.54\% & & &\\
    \midrule
    \multirow{2}{*}{\makebox[0pt][c]{\(\boldsymbol{x}'_{(2)}\)}} & \makebox[0pt][c]{\(S_k\)} & -0.674 & -1.434 & -0.398 & 2.864 & -0.488 & \textbf{3.367} & -0.371 & -0.613 & -1.421 & -0.833 & \multirow{2}{*}{1.057} & \multirow{2}{*}{6} & \multirow{2}{*}{\textcolor{green!70!black}{Successful}} \\
    & \makebox[0pt][c]{\(P_k\)} & 1.01\% &  0.47\% &  1.33\% & 34.74\% & 1.22\% & \textbf{57.45\%} & 1.37\% & 1.07\% & 0.48\% & 0.86\% & & &\\
    \bottomrule
  \end{tabularx}
  }
  \vspace{2pt}
\end{table*}

\begin{figure*}[t]
  \centering
  \includegraphics[width=0.98\linewidth]{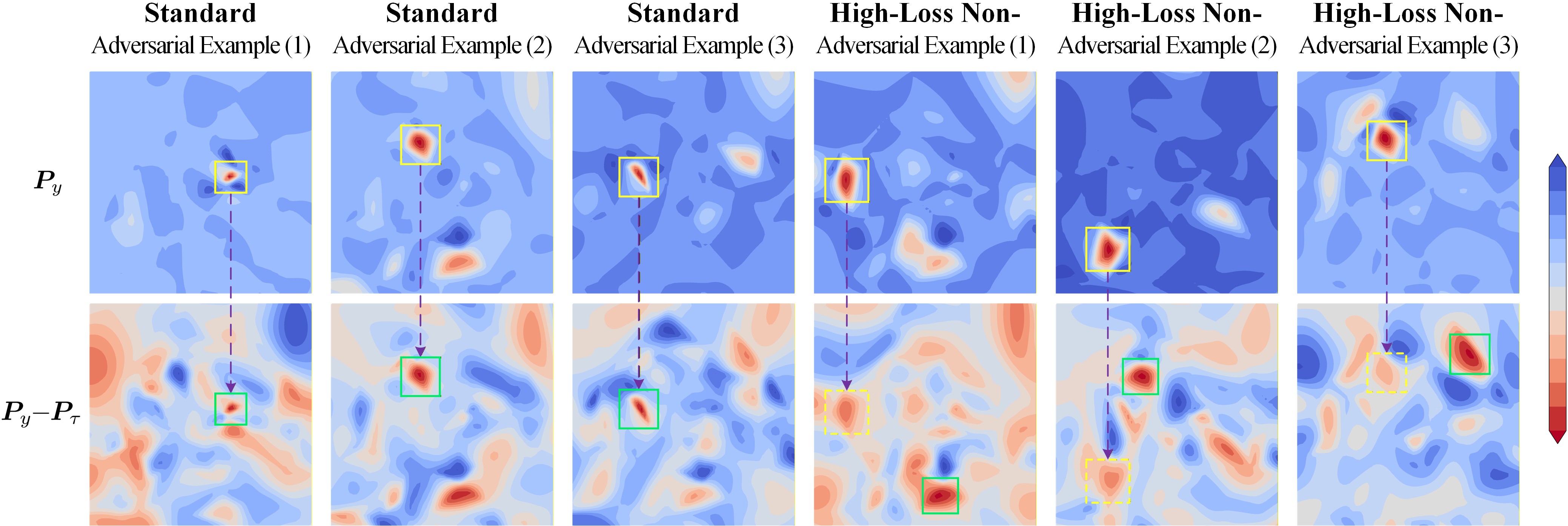}
  \caption{Probability landscapes of standard adversarial examples and high-loss non-adversarial examples.}
  \label{Fig1}
  \vspace{2pt}
\end{figure*}

As illustrated in Section~\ref{Sec2}, attack methods such as FGSM and PGD set the generation objective of adversarial examples as \textbf{minimizing the score (probability) of ground-truth label}, which is achieved by maximizing the CE loss. However, is it truly reasonable?

We argue that this setup is \textbf{unreasonable}. Next, we will elaborate on our viewpoint by discussing an interesting issue termed ``high-loss non-adversarial examples".

For a natural example \(\boldsymbol{x}\) with the ground-truth label \(y=4\), \(\boldsymbol{x}'_{(1)}\) and \(\boldsymbol{x}'_{(2)}\) are two adversarial examples generated under the perturbation budget of \(\bm{\ell}_{\infty}=8/255\). When \(\boldsymbol{x}'_{(1)}\) and \(\boldsymbol{x}'_{(2)}\) are fed into the classification model, their output performances are presented in Table~\ref{Tab1}. Herein, the CE loss is computed via PyTorch's CrossEntropyLoss function. \(\boldsymbol{S}_k\) denotes the output score of the example for label \(k\), i.e., \(f_k(\boldsymbol{x}')\). \(\boldsymbol{P}_k\) represents the normalized probability of the example for label \(k\), which is expressed as follows:

\vspace{-5pt}
\begin{equation}
\boldsymbol{P}_k=\frac{\text{e}^{\boldsymbol{S}_k}}{\sum_{i=1}^K{\text{e}^{\boldsymbol{S}_i}}}.
\label{Equ3}
\end{equation}
\vspace{-6pt}

For example \(\boldsymbol{x}'_{(1)}\), the score and probability at the predicted label ``4" are the highest. Its predicted label matched the ground-truth label, and the attack is thus deemed failed, with a corresponding CE loss of 1.196. For example \(\boldsymbol{x}'_{(2)}\), the score and probability at the predicted label ``6" are the highest. Its predicted label is different from the ground-truth label, and the attack was therefore judged successful, with a corresponding CE loss of 1.057. It can be observed that for example \(\boldsymbol{x}'_{(2)}\), the attack succeeds despite a low CE loss; whereas for example \(\boldsymbol{x}'_{(1)}\), the attack fails even with a higher CE loss. Example \(\boldsymbol{x}'_{(1)}\) corresponds to the ``high-loss non-adversarial example" mentioned previously. For further in-depth empirical analysis of this issue, see Section~\ref{Sec53}.

This issue occurs because minimizing the ground-truth label probability does not guarantee that it falls below the maximum non-ground-truth probability, which is the prerequisite for generating adversarial examples. To elucidate this reason, we analyze the probability landscapes of six examples under perturbation, as visualized in Figure~\ref{Fig1}. The detailed construction procedure of probability landscapes is presented as follows. First, we add diverse random perturbations to natural examples within the predefined perturbation budget. All these perturbations are projected onto a two-dimensional plane via Principal Component Analysis (PCA), and each perturbation is assigned a corresponding two-dimensional coordinate. Afterwards, the classifier makes predictions on the perturbed examples. We further compute the confidence score of the ground-truth class and the probability gap between it and the most probable incorrect class. Finally, we conduct interpolation utilizing these two-dimensional coordinates and probability values to construct two groups of smooth probability distribution graphs, which intuitively demonstrate the variation of prediction probabilities with input perturbations. 

\vspace{3pt}

As shown in Figure~\ref{Fig1}, the upper panels show the ground-truth label probability \(\boldsymbol{P}_y\), with the yellow box marking its minimum region. The lower panels plot the corresponding probability difference (\(\boldsymbol{P}_y-\boldsymbol{P}_{\tau}\), ground-truth label \(y\) minus max non-ground-truth label \(\tau\)), with the green box indicating its minimum. For the first three examples, these minimum regions coincide; thus, PGD successfully generates adversarial examples by minimizing the former. For the latter three, the regions are misaligned. Consequently, PGD's minimization of the ground-truth label probability fails to minimize the difference, yielding ``high-loss non-adversarial examples" instead.

The existence of ``high-loss non-adversarial examples" demonstrates that a low ground-truth label probability and a high CE loss value are not sufficient conditions for generating adversarial examples. Consequently, the generation objectives and loss functions of adversarial examples require further exploration and re-evaluation.

\section{Methodology}
\label{Sec4}

In this section, we reconstruct the generation objective and propose a novel gradient-based attack method to encompass a broader generation of adversarial examples.

\subsection{Reconstruction of Generation Objective}

In attack methods such as PGD, the generation objective is set to ``minimize the ground-truth label probability". This objective tends to produce ``high-loss non-adversarial examples", thereby weakening the attack potency and significantly undermining the validity of target model evaluation.

Inspired by the above discussion, we recognize that both the ground-truth label probability and the non-ground-truth label probabilities must be constrained during generating adversarial examples. To this end, we reconstruct the generation objective of adversarial examples as ``maximizing the difference between the non-ground-truth label probability upper bound and the ground-truth label probability", which aims to achieve successful attacks by constraining all labels. This generation objective can be formulated as follows:

\begin{equation}
\max_{\boldsymbol{x}'} \left\{\boldsymbol{P}_{\tau}-\boldsymbol{P}_y\right\} \quad \text{s.t.} \quad \left\|\boldsymbol{x}' - \boldsymbol{x}\right\|_p \leq \epsilon,
\label{Equ4}
\end{equation}

The generation objective \ref{Equ4} features two optimization directions: ``reducing the ground-truth label probability" and ``increasing the non-ground-truth label probability upper bound". We prove in Section~\ref{SecA} that these two directions are neither mutually independent nor completely consistent, but rather partially correlated. This correlation significantly increases the optimization difficulty, and directly optimizing it as a loss function tends to trap the process in local optima. In the next section, we proposes a novel solution method.

\subsection{Sequential Difference Maximization}

Sequential optimization is a commonly adopted solution method for multi-variable optimization problems. Inspired by this idea, we propose \textbf{Sequential Difference Maximization} (\textbf{SDM}) to solve the adversarial example generation problem formulated in Equation~\ref{Equ4}. SDM is a gradient-based multi-stage iterative method, which decomposes the overall optimization objective into multiple optimization sub-objectives. It performs sequential optimization on these sub-objectives (i.e., the optimal solution of the previous stage is used as the initial solution for the subsequent stage), thereby gradually approaching the optimal solution of the overall optimization objective.

\noindent{\textbf{4.2.1 Overall Structure}}

As shown in Figure~\ref{Fig2}, SDM consists of \(C\) \dashuline{identical} optimization \textbf{cycles}. Each cycle comprises contains \(N\) \uline{different} optimization \textbf{stages}, which are differentiated by their unique sub-objectives and loss functions. These stages are executed in a predetermined sequence. Furthermore, each optimization stage contains \(T\) \dashuline{identical} iteration \textbf{steps}.

For the \(t\)-th step, the adversarial example is expressed as:

\begin{equation}
\resizebox{0.91\hsize}{!}{\(
\boldsymbol{x}_{\text{adv}}^{(t)}=\boldsymbol{x}+\text{Proj}\left(\boldsymbol{x}_{\text{adv}}^{(t-1)}-\boldsymbol{x}+\alpha\cdot\text{sign}(Grad),-\epsilon,\epsilon\right),
\)}
\label{Equ19}
\end{equation}

where \(\alpha\) denotes the step size for adjusting the adversarial perturbation, and \(\text{Proj}(\cdot)\) and \(\text{sign}(\cdot)\) represent the projection function and sign function, respectively. Under the \(\bm{\ell}_{\infty}\)-norm constraint, \(\text{Proj}(\cdot)\) is formulated as:

\begin{figure}[t]
  \centering
  \includegraphics[width=0.96\linewidth]{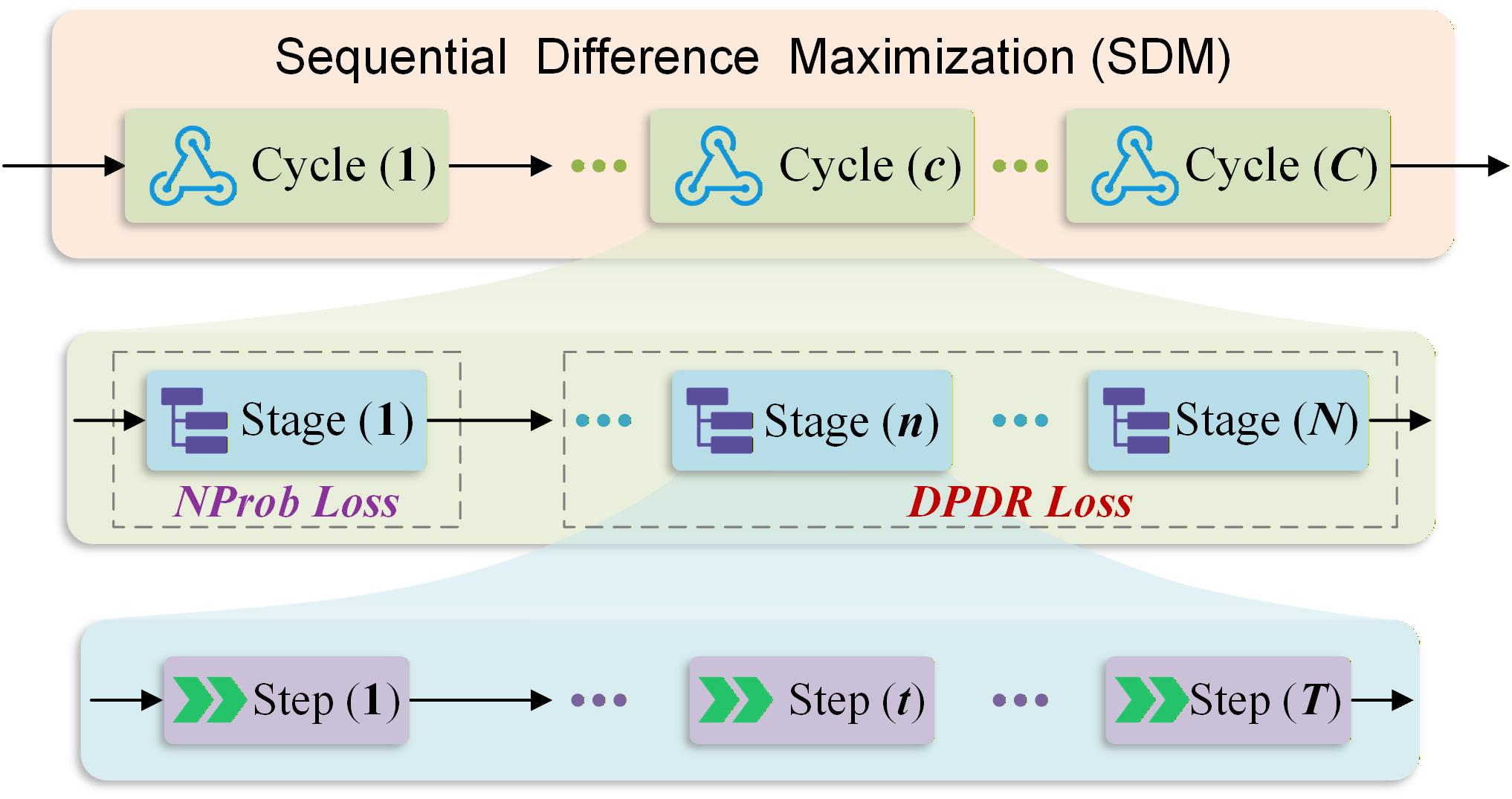}
  \caption{Overall structure of SDM.}
  \label{Fig2}
\end{figure}

\begin{equation}
\text{Proj}\left(\boldsymbol{x},a,b\right)=\max\left(a,\min\left(\boldsymbol{x},b\right)\right).
\label{Equ20}
\end{equation}

\(Grad\) denotes the gradient of the loss between the output probability distribution and the ground-truth label with respect to the adversarial example, which is expressed as:

\begin{equation}
Grad=\nabla_{\boldsymbol{x}_{\text{adv}}^{(t-1)}}L\left(\boldsymbol{P},y\right),
\label{Equ21}
\end{equation}

where the loss function \(L\) varies across each optimization stage, which will be elaborated on in subsequent subsections. In addition, the projection constraint method under the \(\bm{\ell}_2\)-norm is detailed in Section~\ref{SecB}.

\noindent{\textbf{4.2.2 Sub-Objective and Loss Function for Initial Stage}}

SDM sets the optimization sub-objective of the initial optimization stage as ``reducing the ground-truth label probability", and accordingly defines the negative probability loss function \(L_\text{NProb}\) as follows:

\vspace{-3.5pt}
\begin{equation}
L_\text{NProb}=-\boldsymbol{P}_y.
\label{Equ22}
\end{equation}

Compared with directly optimizing the non-ground-truth label probability upper bound, prioritizing the optimization of the ground-truth label probability can effectively compress the solution space, thereby facilitating subsequent optimization steps.

\noindent{\textbf{4.2.3 Sub-Objectives and Loss Functions for Subsequent Stages}}

For \(n\)-th (\(n\geq2\)) stage, SDM sets its optimization sub-objectives to two aspects: (1) maintaining the ground-truth label probability at a low value; (2) reducing the \(n\)-th label probability after descending sorting, so as to elevate the non-ground-truth label probability upper bound. To achieve these objectives, this paper proposes a novel loss function termed \textbf{Directional Probability Difference Ratio} (\textbf{DPDR}), whose mathematical expression is given as follows:

\vspace{-7pt}
\begin{equation}
L_{\text{DPDR}}^{\left(n\right)}=\frac{\boldsymbol{P}_{\tau}-\boldsymbol{P}_y}{\phi-\text{sign}(\boldsymbol{P}_{\tau}-\boldsymbol{P}_y)\cdot(\boldsymbol{P}_{\tau}-\grave{\boldsymbol{P}}_n-\phi)+\zeta}.
\label{Equ23}
\end{equation}

In Equation~\ref{Equ23}, \(\zeta\) is a numerical stability parameter with a default value of \(10^{-10}\), which can avoid numerical anomalies caused by the denominator approaching zero. \(\grave{\boldsymbol{P}}\) represents the result of sorting the probability distribution \(\boldsymbol{P}\) in descending order, which satisfies \(\grave{\boldsymbol{P}}_1\geq\grave{\boldsymbol{P}}_2\geq\cdots\geq\grave{\boldsymbol{P}}_K\geq{0}\). Specifically,  \(\grave{\boldsymbol{P}}_1\) is the maximum value among the probabilities of all labels, \(\grave{\boldsymbol{P}}_2\) is the second largest value, and so on. In addition, \(\phi\) serves as a bias term to ensure that the denominator in the above equation is always greater than 0, which is expressed as:

\begin{equation}
\phi=0.5\times\max\left(\boldsymbol{P}_{\tau}-\grave{\boldsymbol{P}}_n\right).
\label{Equ24}
\end{equation}

The optimization mechanism of the DPDR loss function is described as follows:

(1) When the adversarial attack fails, i.e., the non-ground-truth label probability upper bound is less than the ground-truth label probability (\(\boldsymbol{P}_\tau<\boldsymbol{P}_y\)), the denominator of the DPDR loss \(L_{\text{DPDR}}^{\left(n\right)}\) can be simplified to \(\boldsymbol{P}_{\tau}-\grave{\boldsymbol{P}}_n-\phi\). In this case, the numerator of \(L_{\text{DPDR}}^{\left(n\right)}\) is always negative while its denominator is always positive. Increasing the value of \(L_{\text{DPDR}}^{\left(n\right)}\) will lead to simultaneous increases in the values of both its numerator and denominator, which ultimately achieves the effect of compressing the \(n\)-th largest label probability \(\grave{\boldsymbol{P}}_n\) to increase the largest non-ground-truth label probability (the non-ground-truth label probability upper bound \(\boldsymbol{P}_\tau\)).

(2) When the adversarial attack succeeds, i.e., the non-ground-truth label probability upper bound is greater than the ground-truth label probability (\(\boldsymbol{P}_\tau>\boldsymbol{P}_y\)), the denominator of the DPDR loss \(L_{\text{DPDR}}^{\left(n\right)}\) can be simplified to \(\max(\boldsymbol{P}_{\tau}-\grave{\boldsymbol{P}}_n)-(\boldsymbol{P}_{\tau}-\grave{\boldsymbol{P}}_n)+\zeta\). In this case, both the numerator and denominator of \(L_{\text{DPDR}}^{\left(n\right)}\) are always positive. Increasing the value of \(L_{\text{DPDR}}^{\left(n\right)}\) will raise the value of its numerator and reduce the value of its denominator, which ultimately still achieves the effect of compressing the \(n\)-th largest label probability \(\grave{\boldsymbol{P}}_n\) to increase the non-ground-truth label probability upper bound \(\boldsymbol{P}_\tau\).

It can be observed that, due to the adoption of the \(\text{sign}(\cdot)\) function in the DPDR loss, it is guaranteed to play a role in compressing the \(n\)-th largest label probability regardless of whether the attack succeeds or not. Through this approach, the DPDR loss gradually increases the non-ground-truth label probability upper bound, thereby achieving consistency and stability of the optimization direction in the process of adversarial example generation.

\begin{algorithm}[t]
	\caption{Detailed procedure of the SDM algorithm under \(\bm{\ell}_\infty\)-norm constraint.}
	\label{Alg1}
	\begin{algorithmic}[1]
            \REQUIRE A natural example \(\boldsymbol{x}\) with the ground-truth label \(y\), a perturbation budget \(\epsilon\) and step size \(\alpha\), the number of optimization cycles \( C \), the number of optimization stages \( N \), and the number of iteration steps \( T \).
		\ENSURE The adversarial example \(\boldsymbol{x}'\).
            \STATE Set the initial value of the adversarial example as \(\boldsymbol{x}'_0=\boldsymbol{x}\);
            \FOR{\(c=1\) to \(C\)}
                \FOR{\(n=1\) to \(N\)}
                    \FOR{\(t=1\) to \(T\)}
                        \IF{\(n\) \text{is} \(1\)}
                            \STATE Set the loss function as \(L=L_\text{NProb}\);
                        \ELSE
                            \STATE Set the loss function as \(L=L_\text{DPDR}^{(n)}\);
                        \ENDIF
                        \STATE Obtain the probability distribution \(\boldsymbol{P}\) of the adversarial example \(\boldsymbol{x}'_{t-1}\);
                        \STATE Calculate the loss \(L\left(\boldsymbol{P},y\right)\);
                        \STATE Compute the adversarial example's gradient \(Grad=\nabla_{\boldsymbol{x}'_{t-1}}L(\boldsymbol{P},y)\) of loss;
                        \STATE Update the adversarial example \(\boldsymbol{x}'_t=\boldsymbol{x}+\text{Proj}\left(\boldsymbol{x}'_{t-1}-\boldsymbol{x}+\alpha\cdot\text{sign}(Grad),-\epsilon,\epsilon\right)\);
                    \ENDFOR
                    \STATE Set the initial value of the adversarial example as \(\boldsymbol{x}'_0=\boldsymbol{x}'_T\).
                \ENDFOR
            \ENDFOR
	\end{algorithmic}
\end{algorithm}

\vspace{4pt}
\noindent{\textbf{4.2.4 Detailed Algorithm Flow}}

This section summarizes the implementation process of the SDM algorithm as follows:

(1) In the initial optimization stage (Optimization Stage 1), the negative probability loss \(L_\text{NProb}\) is adopted as the loss function, and gradient ascent is performed through multiple iterative steps to find the optimal solution for Stage 1.

(2) Optimization Stage 2 takes the optimal solution of Optimization Stage 1 as the initial solution, adopts \(L_{\text{DPDR}}^{\left(2\right)}\) as the loss function, and repeats the same multi-step iterative process as in Optimization Stage 1 to find the optimal solution for Stage 2. Subsequently, Optimization Stage 3 takes the optimal solution of Optimization Stage 2 as the initial solution, adopts \(L_{\text{DPDR}}^{\left(3\right)}\) as the loss function, and continues the aforementioned iterative steps. This process is repeated until Optimization Stage \(N\) is completed. In this process, SDM gradually reduces the probability values of the 2nd, 3rd, ..., and ultimately the \(N\)-th largest label probability through the DPDR loss function, thereby continuously elevating the non-ground-truth label probability upper bound.

(3) Optimization Stages 1 to \(N\) are executed sequentially to form a complete optimization cycle. After completing one optimization cycle, a new optimization cycle is initiated to mitigate systematic blind spots in the optimization process.

We adapts the specific process of the SDM algorithm to two norm constraint scenarios, namely the \(\bm{\ell}_\infty\)- and \(\bm{\ell}_2\)-norm constraints. Among them, the detailed implementation process under the \(\bm{\ell}_\infty\)-norm constraint is referred to Algorithm~\ref{Alg1}. The procedure of the SDM algorithm under the \(\bm{\ell}_2\)-norm constraint is similar to that under the \(\bm{\ell}_\infty\)-norm constraint, with the only difference lying in the method of projecting the adversarial example back to the perturbation domain. 

\vspace{4pt}
\section{Experiments}
\label{Sec5}

\subsection{Experimental Setup}

\vspace{1.1pt}
\scalebox{0.97}[1]{\textbf{Models and Datasets.} This study aims to generate} adversarial examples in white-box scenarios. ResNet-34 \cite{HeDee16} and WideResNet-28-10 \cite{Zag16} are adopted as the backbone models. CIFAR-10, CIFAR-100 \cite{Kri09} and Mini-ImageNet \cite{Vin16} are used as the datasets. 

\vspace{1.1pt}
\textbf{Baseline Defenses.} Defense methods under attack include Vanilla Adversarial Training (VAT) \cite{Mad18}, MART \cite{Wan20}, HAT \cite{Rad22}, and LOAT \cite{Yin24}. All utilize PGD \cite{Mad18} with a perturbation budget of \(\boldsymbol{\bm{\ell}}_\infty=8/255\), step size of \(\boldsymbol{\bm{\ell}}_\infty=2/255\), and 10 iterations to generate adversarial examples for training. Unspecified training settings follow the original configurations.

\textbf{Baseline Attacks.} Attack methods competing with the proposed SDM include PGD \cite{Mad18}, C\&W \cite{Car17}, APGD \cite{Cro20}, and AA \cite{Cro20}. Furthermore, to ensure fair comparisons, we adopt the same strategy as PGD to constrain the perturbation budget of C\&W within the predefined range. The specific attack parameters will be elaborated on in subsequent evaluation experiments.

\begin{table}[t]
\caption{Detailed configuration of total steps in SDM.}
\label{Tab2}
  \centering
  \newcolumntype{C}{>{\centering}X}
  \newcolumntype{L}[1]{>{\centering\arraybackslash}p{#1}}
  \scalebox{0.9}{
  \begin{tabularx}{1.05\linewidth}{C|*{7}{L{0.53cm}}}
    \toprule
    \makebox[0pt][c]{Total Steps} & \makebox[0pt][c]{10} & \makebox[0pt][c]{20} & \makebox[0pt][c]{50} & \makebox[0pt][c]{100} & \makebox[0pt][c]{200} & \makebox[0pt][c]{500} & \makebox[0pt][c]{1000}\\
    \midrule
    \makebox[0pt][c]{Cycles} & \makebox[0pt][c]{1} & \makebox[0pt][c]{1} & \makebox[0pt][c]{2} & \makebox[0pt][c]{2} & \makebox[0pt][c]{4} & \makebox[0pt][c]{4} & \makebox[0pt][c]{5}\\
    \makebox[0pt][c]{Stages} & \makebox[0pt][c]{5} & \makebox[0pt][c]{5} & \makebox[0pt][c]{5} & \makebox[0pt][c]{5} & \makebox[0pt][c]{5} & \makebox[0pt][c]{5} & \makebox[0pt][c]{5}\\
    \makebox[0pt][c]{Steps} & \makebox[0pt][c]{2} & \makebox[0pt][c]{4} & \makebox[0pt][c]{5} & \makebox[0pt][c]{10} & \makebox[0pt][c]{10} & \makebox[0pt][c]{25} & \makebox[0pt][c]{40}\\
    \bottomrule
  \end{tabularx}
  }
\end{table}

\subsection{Determination of Hyperparameter Combinations}

The SDM algorithm involves four hyperparameters, namely the total iteration steps \(Z\), the number of optimization cycles \(C\), the number of optimization stages \(N\), and the number of iterative steps per stage \(T\). The relationship among these four parameters is \(Z=C\times{N}\times{T}\). These hyperparameters are closely correlated with the performance of SDM—an inappropriate selection of hyperparameters will significantly degrade the algorithm's performance. However, the problem of determining their allocation for efficiency maximization is difficult to analyze quantitatively at a theoretical level. For this reason, we have conducted extensive experiments to identify the optimal correspondence between these hyperparameters, as show in Section~\ref{SecC}. All optimal hyperparameter combinations are collated uniformly in Table~\ref{Tab2}. In the subsequent experiments, only the total iteration steps of SDM is specified, and the corresponding specific values of the number of optimization cycles, the number of optimization stages, and the number of iterative steps per stage can be referred to the detailed data in this table.

\subsection{Empirical Analysis of High-Loss Non-Adversarial Examples}
\label{Sec53}

High-loss non-adversarial examples are inputs that fail to fool the target model but yield a higher CE loss than successfully examples. This phenomenon cannot be identified using only PGD, as relative loss levels lack a reference. Only by comparing PGD with other attacks and analyzing examples where PGD fails but other methods succeed can we reveal the existence of this examples. The motivating case in Section~\ref{Sec3} originated from PGD-C\&W comparisons. Now, we designed comprehensive controlled experiments to validate this issue.

We use C\&W, FAB, and Jitter as baseline attacks, and PGD and SDM as methods under evaluation. We trained 6 robust models via VAT on CIFAR-10: ResNet-18, GoogLeNet, VGG-19, DenseNet-121, DPN-26 and ShuffleNetV2-20. We generate adversarial sets \(\boldsymbol{x}_{cw}\), \(\boldsymbol{x}_{pgd}\) and \(\boldsymbol{x}_{sdm}\). For examples where PGD fails but C\&W succeeds (indexed by \(h\)), we compute CE loss differences \(L_1=L(\boldsymbol{x}_{pgd}[h])-L(\boldsymbol{x}_{cw}[h])\) and  \(L_2=L(\boldsymbol{x}_{sdm}[h])-L(\boldsymbol{x}_{cw}[h])\). Results of the first three models are listed in Table~\ref{Tab13}, with consistent findings on the rest models.

The table is interpreted as follows. For the first group ``190/0.105" and ``20/0.046", 190 denotes the number of examples where PGD fails but C\&W succeeds (i.e., \(\boldsymbol{x}_{pgd}[h]\)), and 0.105 is the mean value of \(L_1\); 20 denotes the number of such examples on which SDM fails (i.e., failed examples in \(\boldsymbol{x}_{sdm}[h]\)), and 0.046 is the mean value of \(L_2\). 

The mean values of \(L_1\) are all positive. Combined with their attack failure, these PGD examples correspond to the high-loss non-adversarial examples. Compared with the high \(L_1\) of PGD examples, SDM examples with low \(L_2\) reduce attack failure rate from 100\% to 9.06\%. This shows that SDM effectively mitigates high-loss non-adversarial examples and improves the attack performance. We summarize as follows: (1) High-loss non-adversarial examples are a relative metric requiring comparison between PGD and other attacks. (2) They widely occur across evaluation settings, with an incidence rate of 1.5\%–3\%. (3) SDM reduces their occurrence by 90\%–95\%, substantially alleviating the issue.

\begin{table}[t!]
\caption{Occurrence rate of high-loss non-adversarial examples.}
\label{Tab13}
  \centering
  \newcolumntype{C}{>{\centering}X}
  \newcolumntype{L}[1]{>{\centering\arraybackslash}p{#1}}
  \scalebox{0.88}{
  \begin{tabularx}{1.1\linewidth}{CL{1.1cm}|*{3}{L{1.5cm}}}
    \toprule
    \makebox[0pt][c]{Model} & \makebox[0pt][c]{Attack} & \makebox[0pt][c]{C\&W} & \makebox[0pt][c]{FAB} & \makebox[0pt][c]{Jitter}\\
    \midrule
    \multirow{2}{*}{\makebox[0pt][c]{ResNet-18}} & \makebox[0pt][c]{PGD} & \makebox[0pt][c]{190 / 0.105} & \makebox[0pt][c]{213 / 0.201} & \makebox[0pt][c]{173 / 0.107}\\
    & \makebox[0pt][c]{SDM} & \makebox[0pt][c]{\phantom{0}20 / 0.046} & \makebox[0pt][c]{\phantom{0}19 / 0.142} & \makebox[0pt][c]{\phantom{0}20 / 0.048}\\
    \midrule
    \multirow{2}{*}{\makebox[0pt][c]{GoogLeNet}} & \makebox[0pt][c]{PGD} & \makebox[0pt][c]{192 / 0.097} & \makebox[0pt][c]{212 / 0.182} & \makebox[0pt][c]{180 / 0.106}\\
    & \makebox[0pt][c]{SDM} & \makebox[0pt][c]{\phantom{0}19 / 0.040} & \makebox[0pt][c]{\phantom{0}17 / 0.129} & \makebox[0pt][c]{\phantom{0}17 / 0.050}\\
    \midrule
    \multirow{2}{*}{\makebox[0pt][c]{VGG-19}} & \makebox[0pt][c]{PGD} & \makebox[0pt][c]{202 / 0.111} & \makebox[0pt][c]{215 / 0.202} & \makebox[0pt][c]{192 / 0.113}\\
    & \makebox[0pt][c]{SDM} & \makebox[0pt][c]{\phantom{0}16 / 0.037} & \makebox[0pt][c]{\phantom{0}16 / 0.124} & \makebox[0pt][c]{\phantom{0}16 / 0.038}\\
    \bottomrule
  \end{tabularx}
  }
\end{table}

\subsection{Generational Relationships among Attack Methods}

\begin{figure}[t]
  \centering
  \includegraphics[width=0.9\linewidth]{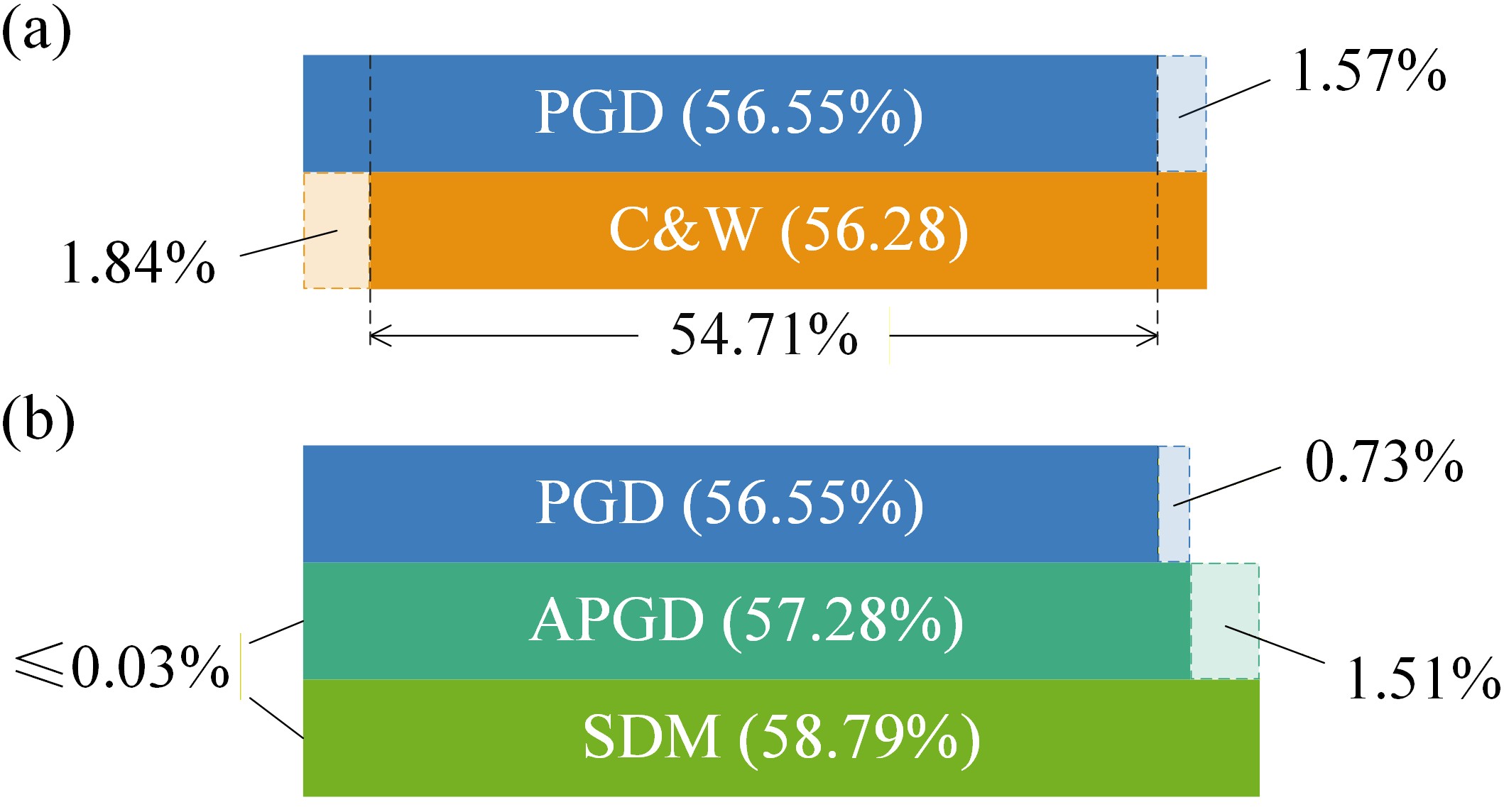}
  \caption{Intersection and difference proportions of adversarial examples generated by various attack methods.}
  \label{Fig3}
\end{figure}

We evaluated various adversarial attacks on CIFAR-10, including PGD, APGD, C\&W, and SDM, under a perturbation budget of 8/255 and a step size of 2/255. PGD, C\&W, and SDM were run for 500 iterations, while APGD used 50 iterations with 10 restarts. All experiments employed a ResNet-34 model defended by VAT. After the attacks, we recorded the successfully attacked examples for each method and analyzed their overlaps and differences, with the results summarized in Figure~\ref{Fig3}.

\begin{table*}[t]
\caption{Attack success rates of various attack methods under \(\bm{\ell}_{\infty}\)-norm constraint.}
\label{Tab3}
\centering
  \newcolumntype{C}{>{\centering}X}
  \newcolumntype{L}[1]{>{\centering\arraybackslash}p{#1}}
  \scalebox{0.88}{
  \begin{tabularx}{1.09\linewidth}{C|L{0.95cm}|*{4}{L{0.76cm}}|*{4}{L{0.76cm}}|*{4}{L{0.76cm}}}
    \toprule
    \multirow{2}{*}{\makebox[0pt][c|]{Model}} & \multirow{2}{*}{\makebox[0pt][c|]{Attack}} & \multicolumn{4}{c|}{CIFAR-10} & \multicolumn{4}{c|}{CIFAR-100} & \multicolumn{4}{c}{Mini-ImageNet}\\
    & & \makebox[0pt][c]{VAT} & \makebox[0pt][c]{MART} & \makebox[0pt][c]{HAT} & \makebox[0pt][c]{LOAT} & \makebox[0pt][c]{VAT} & \makebox[0pt][c]{MART} & \makebox[0pt][c]{HAT} & \makebox[0pt][c]{LOAT} & \makebox[0pt][c]{VAT} & \makebox[0pt][c]{MART} & \makebox[0pt][c]{HAT} & \makebox[0pt][c]{LOAT}\\
    \midrule
    \multirow{4}{*}{\makebox[0pt][c]{ResNet-34}} & \makebox[0pt][c]{PGD} & 56.55 & 48.47 & 46.91 & 45.10 & 77.23 & 72.53 & 70.55 & 68.96 & 78.34  & 75.97 & 73.49 & 71.14\\
    & \makebox[0pt][c]{C\&W} & 56.28 & 53.70 & 50.63 & 49.28 & 78.06 & 77.11 & 75.96 & 72.16 & 80.47 & 78.96 & 76.80 & 74.42 \\
    & \makebox[0pt][c]{APGD} & 57.28 & 49.37 & 47.66 & 45.96 & 78.20 & 73.20 & 71.13 & 69.56 & 79.15 & 77.57 & 74.20 & 71.85\\
    & \makebox[0pt][c]{SDM} & 58.79 & 54.43 & 52.92 & 51.04 & 80.63 & 77.89 & 76.76 & 74.17 & 83.03 & 81.46 & 79.05 & 77.32\\
    \midrule
    \multirow{4}{*}{\makebox[0pt][c]{WideResNet-28-10}} & \makebox[0pt][c]{PGD} & 57.88 & 51.50 & 48.79 & 46.85 & 77.97 & 73.37 & 71.74 & 69.71 & 80.80 & 73.52 & 71.10 & 70.95\\
    & \makebox[0pt][c]{C\&W} & 58.14 & 57.70 & 53.96 & 50.93 & 79.17 & 77.76 & 74.39 & 73.03 & 81.74 & 78.88 & 74.82 & 73.03\\
    & \makebox[0pt][c]{APGD} & 58.36 & 52.46 & 49.62 & 47.54 & 78.76 & 73.91 & 72.30 & 70.28 & 81.52 & 73.89 & 71.82 & 71.69\\
    & \makebox[0pt][c]{SDM} & 60.31 & 58.59 & 54.66 & 52.68 & 81.47 & 78.44 & 76.66 & 74.28 & 84.40 & 79.97 & 76.69 & 75.44\\
    \bottomrule
  \end{tabularx}
  }
\end{table*}

\begin{table*}[t]
\caption{Attack success rates of various attack methods under \(\bm{\ell}_2\)-norm constraint.}
\label{Tab4}
  \centering
  \newcolumntype{C}{>{\centering}X}
  \newcolumntype{L}[1]{>{\centering\arraybackslash}p{#1}}
  \scalebox{0.88}{
  \begin{tabularx}{1.09\linewidth}{C|L{0.95cm}|*{4}{L{0.76cm}}|*{4}{L{0.76cm}}|*{4}{L{0.76cm}}}
    \toprule
    \multirow{2}{*}{\makebox[0pt][c|]{Model}} & \multirow{2}{*}{\makebox[0pt][c|]{Attack}} & \multicolumn{4}{c|}{CIFAR-10} & \multicolumn{4}{c|}{CIFAR-100} & \multicolumn{4}{c}{Mini-ImageNet}\\
    & & \makebox[0pt][c]{VAT} & \makebox[0pt][c]{MART} & \makebox[0pt][c]{HAT} & \makebox[0pt][c]{LOAT} & \makebox[0pt][c]{VAT} & \makebox[0pt][c]{MART} & \makebox[0pt][c]{HAT} & \makebox[0pt][c]{LOAT} & \makebox[0pt][c]{VAT} & \makebox[0pt][c]{MART} & \makebox[0pt][c]{HAT} & \makebox[0pt][c]{LOAT}\\
    \midrule
    \multirow{4}{*}{\makebox[0pt][c]{ResNet-34}} & \makebox[0pt][c]{PGD} & 71.12 & 60.42 & 57.38 & 55.65 & 81.83 & 77.08 & 74.16 & 72.22 & 75.51 & 67.96 & 65.25 & 63.25\\
    & \makebox[0pt][c]{C\&W} & 70.26 & 48.60 & 45.77 & 43.71 & 79.71 & 72.11 & 69.12 & 66.75 & 53.22 & 48.55 & 46.81 & 45.62\\
    & \makebox[0pt][c]{APGD} & 71.95 & 62.05 & 57.92 & 56.08 & 82.81 & 78.18 & 74.79 & 72.85 & 76.52 & 68.71 & 65.93 & 63.83\\
    & \makebox[0pt][c]{SDM} & 72.64 & 68.02 & 65.80 & 63.75 & 84.50 & 82.38 & 79.50 & 77.26 & 80.24 & 77.14 & 74.24 & 71.47\\
    \midrule
    \multirow{4}{*}{\makebox[0pt][c]{WideResNet-28-10}} & \makebox[0pt][c]{PGD} & 68.79 & 57.93 & 54.19 & 52.28 & 79.72 & 75.32 & 72.96 & 70.51 & 77.52 & 72.96 & 69.93 & 67.35\\
    & \makebox[0pt][c]{C\&W} & 67.71 & 45.33 & 42.49 & 41.68 & 78.68 & 70.61 & 67.34 & 65.70 & 65.13 & 58.96 & 56.35 & 54.83\\
    & \makebox[0pt][c]{APGD} & 69.53 & 58.99 & 54.89 & 52.76 & 80.60 & 76.08 & 73.74 & 71.24 & 78.26 & 73.75 & 70.65 & 67.93\\
    & \makebox[0pt][c]{SDM} & 70.64 & 65.27 & 62.45 & 60.66 & 82.95 & 80.45 & 78.59 & 75.89 & 81.06 & 79.03 & 76.20 & 74.59\\
    \bottomrule
  \end{tabularx}
  }
\end{table*}

As shown in Figure~\ref{Fig3}(a), the intersection between PGD and C\&W adversarial examples accounts for 54.71\% of the dataset. The set difference of PGD relative to C\&W is 1.84\%, and that of C\&W relative to PGD is 1.57\%. PGD and the C\&W attack employ divergent strategies—iterative gradient ascent versus direct perturbation minimization—yet the results demonstrate that they represent distinct technical pathways, with no significant generational gap or clear performance advantage observed between them.

In Figure~\ref{Fig3}(b), the set difference of PGD relative to APGD is nearly empty (less than 0.03\% of the dataset), while the difference of APGD relative to PGD occupies 0.73\%. This suggests APGD identifies blind spots of PGD attacks, making it a new generation of gradient-based methods. Furthermore, the set difference of SDM relative to APGD is 1.51\%, and relative to PGD it reaches 2.24\%. This demonstrates that SDM constitutes an even more advanced gradient-based approach, capable of producing more robust and threatening adversarial examples. Compared to PGD and APGD, SDM exhibits superior attack performance. Additionally, the average CE loss of examples in SDM’s exclusive set relative to PGD is 1.03, significantly lower than the corresponding value of 1.19 in PGD. This confirms that the latter are ``high-loss non-adversarial examples" as defined in this study, and proving that SDM effectively mitigates this issue.

\subsection{Attack Performance Against Different Defenses}

To verify the attack performance of SDM against diverse defense methods, we design corresponding evaluation experiments under \(\bm{\ell}_{\infty}\)- and \(\bm{\ell}_2\)-norm constraints. Under the \(\bm{\ell}_{\infty}\)-norm constraint, the adversarial perturbation budget and perturbation step size for the evaluated methods (PGD, C\&W, APGD and SDM) are set to 8/255 and 2/255, respectively. The number of restarts for APGD is set to 10, with 50 iterative steps executed before each restart. The iterative steps of the other methods is 500. The results of the evaluation experiments using ResNet-34 and WideResNet-28-10 as target models under the \(\bm{\ell}_{\infty}\)-norm constraint are presented in Table~\ref{Tab3}. Under the \(\bm{\ell}_2\)-norm constraint, the perturbation budget and perturbation step size for PGD, C\&W, APGD and SDM are set to 1.0 and 0.2, respectively, while the iterative steps remains consistent with the experimental setup under the \(\bm{\ell}_{\infty}\)-norm constraint. The evaluation results under the \(\bm{\ell}_2\)-norm constraint are shown in Table~\ref{Tab4}.

\begin{table*}[t]
\caption{Attack success rates of various attack methods under different perturbation budgets.}
\label{Tab5}
  \centering
  \newcolumntype{C}{>{\centering}X}
  \newcolumntype{L}[1]{>{\centering\arraybackslash}p{#1}}
  \scalebox{0.93}{
  \begin{tabularx}{1.02\linewidth}{C|L{1.2cm}|*{6}{L{0.72cm}}|*{6}{L{0.72cm}}}
    \toprule
    \multirow{2}{*}{\makebox[0pt][c|]{Dataset}} & \multirow{2}{*}{\makebox[0pt][c|]{Attack}} & \multicolumn{6}{c|}{\(\bm{\ell}_{\infty}\)} & \multicolumn{6}{c}{\(\bm{\ell}_2\)}\\
    & & \(\tfrac{4}{255}\) & \(\tfrac{8}{255}\) & \(\tfrac{12}{255}\) & \(\tfrac{16}{255}\) & \(\tfrac{20}{255}\) & \(\tfrac{24}{255}\) & 0.5 & 1.0 & 1.5 & 2.0 & 2.5 & 3.0\\
    \midrule
    \multirow{4}{*}{\makebox[0pt][c|]{CIFAR-10}} & \makebox[0pt][c]{PGD} & 31.53 & 56.55 & 78.79 & 92.12 & 97.78 & 99.43 & 36.51 & 71.12 & 92.24 & 98.70 & 99.87 & 99.99\\
    & \makebox[0pt][c]{C\&W} & 32.59 & 56.28 & 77.33 & 90.89 & 97.14 & 99.25 & 37.94 & 70.26 & 89.11 & 94.27 & 94.73 & 94.74\\
    & \makebox[0pt][c]{APGD} & 31.81 & 57.28 & 79.79 & 92.93 & 98.27 & 99.60 & 36.86 & 71.95 & 93.12 & 98.94 & 99.93 & 99.99\\
    & \makebox[0pt][c]{SDM} & 33.21 & 58.79 & 80.16 & 92.63 & 97.92 & 99.46 & 38.17 & 72.64 & 92.59 & 98.72 & 99.85 & 99.98\\
    \midrule
    \multirow{4}{*}{\makebox[0pt][c|]{CIFAR-100}} & \makebox[0pt][c]{PGD} & 59.08 & 77.23 & 89.59 & 95.45 & 98.25 & 99.30 & 60.78 & 81.83 & 93.97 & 98.27 & 99.56 & 99.92\\
    & \makebox[0pt][c]{C\&W} & 61.63 & 78.06 & 88.91 & 94.94 & 97.92 & 99.16 & 63.18 & 79.71 & 86.91 & 88.89 & 89.03 & 89.03\\
    & \makebox[0pt][c]{APGD} & 59.61 & 78.20 & 90.40 & 96.02 & 98.61 & 99.46 & 61.20 & 82.81 & 94.79 & 98.74 & 99.71 & 99.96\\
    & \makebox[0pt][c]{SDM} & 62.87 & 80.63 & 91.46 & 96.40 & 98.59 & 99.44 & 64.29 & 84.50 & 94.94 & 98.66 & 99.61 & 99.92\\
    \bottomrule
  \end{tabularx}
  }
\end{table*}

It can be observed that in the evaluation experiments under the \(\bm{\ell}_{\infty}\)-norm constraint, SDM exhibits a significant performance advantage. When ResNet-34 is adopted as the target model, the average attack success rates of SDM on the CIFAR-10, CIFAR-100 and Mini-ImageNet reach 54.30\%, 77.36\% and 80.22\%, respectively—surpassing the commonly used PGD method by 5.04\%, 5.05\% and 5.48\%, and outperforming APGD by 4.23\%, 4.34\% and 4.52\%, respectively. When WideResNet-28-10 serves as the target model, the average attack success rates of SDM on the CIFAR-10, CIFAR-100 and Mini-ImageNet are 56.56\%, 77.71\% and 79.08\%, respectively, which are 5.31\%, 4.52\% and 4.99\% higher than those of PGD, and 4.57\%, 3.90\% and 4.35\% higher than those of APGD, respectively.

In the evaluations under the \(\bm{\ell}_2\)-norm constraint, SDM continues to demonstrate superior attack performance. When ResNet-34 is the target model, the average attack success rate of SDM on all datasets is 74.75\%—exceeding PGD by 6.26\%, and outperforming APGD by 5.44\%. When WideResNet-28-10 as the target model, the average attack success rate of SDM is 73.98\%, which is 5.69\% higher than that of PGD, and 4.95\% higher than that of APGD.

The experiments above demonstrate that SDM yields significant advantages over the PGD, C\&W, and APGD. Furthermore, in Section~\ref{SecD}, we evaluate the performance of SDM against models with diverse architectures on the ImageNet-1K \cite{LL009} dataset. In Section~\ref{SecE}, we show that SDM achieves performance comparable to that of the integrated method Auto-Attack (AA), and their combination leads to stronger results. Section~\ref{SecF} further verifies the high robustness of the adversarial examples generated by SDM, which maintain the highest attack success rate under various interferences such as mirroring, translation, and noise.

\vspace{3pt}
\subsection{Attack Performance Under Various Perturbation Budgets}
\vspace{1pt}

In this section, systematic adversarial attack experiments are conducted under diverse perturbation budgets to comprehensively and quantitatively evaluate the attack performance of the proposed SDM method. Specifically, we adopt VAT as the baseline defense and PGD, C\&W and APGD as the baseline attacks, and carry out \(\bm{\ell}_{\infty}\)- and \(\bm{\ell}_2\)-norm attack experiments on the CIFAR-10 and CIFAR-100 datasets, respectively. The perturbation ranges for the \(\bm{\ell}_{\infty}\)- and \(\bm{\ell}_2\)-norm are set to \((4/255,24/255)\) and \((0.5,3.0)\), respectively. The attack performance of different attack methods is presented in Table~\ref{Tab5}. It can be observed that the proposed SDM achieves the highest attack performance under all adversarial perturbation budgets in comparison with the baseline attack methods. Under the \(\bm{\ell}_{\infty}\)-norm constraint, the attack success rate of SDM is 1.37\% and 0.80\% higher than those of PGD and APGD, respectively; under the \(\bm{\ell}_2\)-norm constraint, the attack success rate of SDM is 0.93\% and 0.49\% higher than those of PGD and APGD, respectively.

\subsection{Cost-Effectiveness Analysis of Attack Methods}
\label{Sec57}

\begin{figure}[t]
  \centering
  \includegraphics[width=0.9\linewidth]{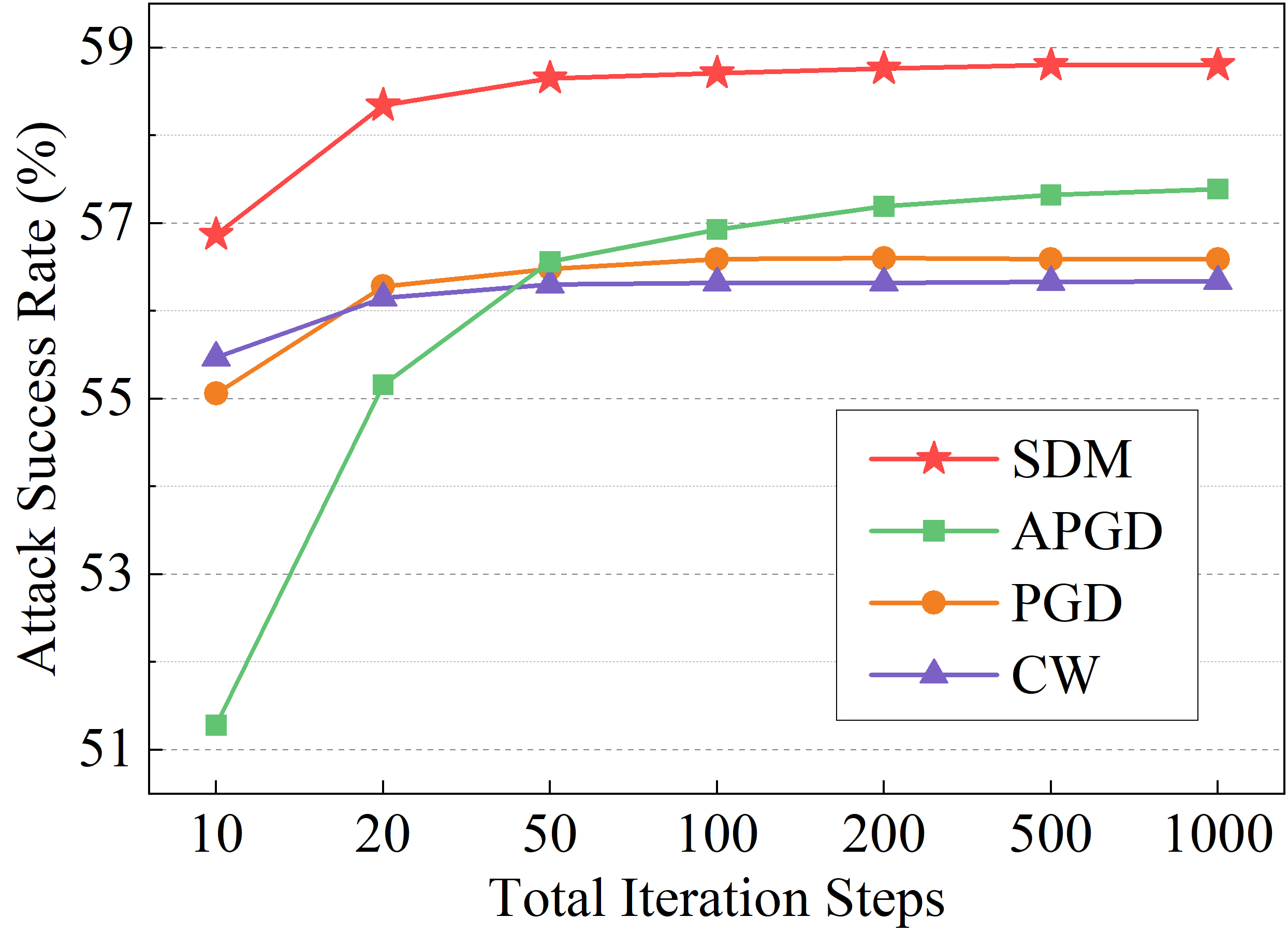}
  \caption{Cost-effectiveness comparison of various attack methods.}
  \label{Fig4}
  \vspace{2.5pt}
\end{figure}

To evaluate the cost-effectiveness of different attack methods, we conduct experiments under varying iteration steps. The experiments employ CIFAR-10 as the dataset, VAT as the baseline defense, and a ResNet-34 as the target model. Under the \(\bm{\ell}_{\infty}\)-norm constraint, the perturbation budget and step size for PGD, C\&W, APGD, and SDM are set to 8/255 and 2/255, respectively. The total number of iterations is set to 10, 20, 50, 100, 200, 500, and 1000. For APGD, the total iteration count is the product of its number of restarts and steps per restart; the number of restarts is set to 5 for total iterations \(\leq\) 100 and 10 otherwise. The attack performance under these settings is shown in Figure~\ref{Fig4}. 

The aforementioned attacks achieve comparable single-run costs (see Section~\ref{SecG}), and their (total) computational cost scales almost linearly with the number of iterations. In Figure~\ref{Fig4}, APGD exhibits the steepest performance growth curve, with its effectiveness stabilizing only after 500 iterations. Consequently, when computational cost is severely constrained (e.g., total iterations \(\leq\) 20), its performance will be significantly inadequate. In contrast, SDM achieves superior performance across all iteration counts, demonstrating a clear advantage. This indicates that SDM offers the highest attack cost-effectiveness, making it highly suitable for scenarios with limited computational resources.

\section{Limitation}
\label{Sec6}

The main limitation of SDM lies in its intricate hyperparameter configuration mechanism. Currently, the optimal assignment of core hyperparameters including total iterations, cycles, stages and step numbers relies heavily on empirical experience, which lacks rigorous theoretical basis and quantitative mathematical interpretation. Although reliable reference values are summarized and provided in Table~\ref{Tab2}, manual parameter calibration and adaptive adjustment may still be required when deploying SDM to novel architectures or datasets. To effectively mitigate this limitation, we recommend adopting the offered reference values as baseline initial configurations for transfer adversarial tasks, and further combining them with Bayesian optimization and advanced auto-tuning algorithms to achieve intelligent, automatic hyperparameter searching and optimization.

\section{Conclusion}
\label{Sec7}

In this paper, we propose a novel gradient-based adversarial attack method termed SDM to alleviate the issue of ``high-loss non-adversarial examples" that restricts the performance of conventional attack methods. Compared with existing adversarial example generation approaches, SDM constructs a more reasonable and challenging optimization objective. To effectively optimize this well-designed objective, we design a three-layer ``cycle-stage-step" iterative optimization framework based on the concept of sequential optimization. This framework progressively approximates the global optimal solution by cyclically imposing and optimizing negative probability loss and DPDR loss throughout predefined optimization stages. Sufficient experimental results verify the superior effectiveness and practicality of the proposed SDM, which consistently outperforms prior state-of-the-art methods in both comprehensive attack performance and computational cost-effectiveness.

\clearpage

\section*{Acknowledgements}

This work was supported in part by the National Key Research and Development Program of China (No. 2024YFB2907202).

\section*{Impact Statement}

This study presents a novel white-box adversarial attack method. It achieves state-of-the-art success rates across various architectures and large-scale datasets, while maintaining computational efficiency comparable to prior approaches. Its positive impact lies in addressing a key limitation of existing attack methods, thereby establishing a more reliable benchmark for evaluating defenses, guiding the design of next-generation defense mechanisms, and ultimately advancing the development of truly secure AI systems. While the proposed method could be misused by attackers and poses a threat to existing defensive measures, this research will be conducted transparently and responsibly to maximize its positive contribution to AI security and minimize potential negative consequences.

\normalem
\bibliography{main}
\bibliographystyle{icml2026}
\clearpage
\newpage

\appendix

\section{Proof of the Relationship Between the Two Optimization Directions}
\label{SecA}

It is known that the probability distribution \(\boldsymbol{P}\) of a example can be expressed according to Equation~\ref{Equ3} as follows:

\begin{equation}
\begin{split}
\boldsymbol{P}&=\left(\boldsymbol{P}_1, \dots, \boldsymbol{P}_{y}, \dots, \boldsymbol{P}_{\tau}, \dots, \boldsymbol{P}_{K} \right)\\
&=\frac{1}{\sum_{i=1}^{K}{\text{e}^{\boldsymbol{S}_i}}}\left(\text{e}^{\boldsymbol{S}_1}, \dots, \text{e}^{\boldsymbol{S}_y}, \dots, \text{e}^{\boldsymbol{S}_{\tau}}, \dots, \text{e}^{\boldsymbol{S}_K}\right)
\end{split}
\label{Equ5}
\end{equation}

The gradient of \(\boldsymbol{P}\) along the optimization direction of ``reducing the ground-truth label probability" can be represented in the form of a partial derivative as:

\begin{equation}
\nabla_{-\boldsymbol{P}_y}\boldsymbol{P}=-\frac{\partial \boldsymbol{P}}{\partial \boldsymbol{S}_y}\cdot\frac{\text{d}\boldsymbol{S}_y}{\text{d}\boldsymbol{P}_y}
\label{Equ6}
\end{equation}

Through calculation, the first term in Equation~\ref{Equ5} can be expressed as:

\begin{equation}
\frac{\partial \boldsymbol{P}}{\partial \boldsymbol{S}_y}=-\boldsymbol{P}_y\left(\boldsymbol{P}_1, \dots, \boldsymbol{P}_{y}-1, \dots, \boldsymbol{P}_{\tau}, \dots, \boldsymbol{P}_{K} \right)
\label{Equ7}
\end{equation}

The second term in Equation~\ref{Equ5} can be expressed as:

\begin{equation}
\frac{\text{d}\boldsymbol{S}_y}{\text{d}\boldsymbol{P}_y}=\frac{1}{\boldsymbol{P}_y\left(1-\boldsymbol{P}_y\right)}
\label{Equ8}
\end{equation}

Since, the simplified result of Equation~\ref{Equ5} is derived as:

\begin{equation}
\nabla_{-\boldsymbol{P}_y}\boldsymbol{P}=\frac{1}{1-\boldsymbol{P}_y}\left(\boldsymbol{P}_1, \dots, \boldsymbol{P}_{y}-1, \dots, \boldsymbol{P}_{\tau}, \dots, \boldsymbol{P}_{K} \right)
\label{Equ11}
\end{equation}

Similarly, the gradient of \(\boldsymbol{P}\) along the optimization direction of ``increasing the non-ground-truth label probability upper bound" can be represented in the form of a partial derivative as:

\begin{equation}
\nabla_{\boldsymbol{P}_{\tau}}\boldsymbol{P}=\frac{\partial \boldsymbol{P}}{\partial \boldsymbol{S}_{\tau}}\cdot\frac{\text{d}\boldsymbol{S}_{\tau}}{\text{d}\boldsymbol{P}_{\tau}}
\label{Equ12}
\end{equation}

Through calculation, the first term in Equation~\ref{Equ12} can be expressed as:

\begin{equation}
\frac{\partial \boldsymbol{P}}{\partial \boldsymbol{S}_{\tau}}=-\boldsymbol{P}_{\tau}\left(\boldsymbol{P}_1, \dots, \boldsymbol{P}_{y}, \dots, \boldsymbol{P}_{\tau}-1, \dots, \boldsymbol{P}_{K} \right)
\label{Equ13}
\end{equation}

The second term in Equation~\ref{Equ12} can be expressed as:

\begin{equation}
\frac{\text{d}\boldsymbol{S}_{\tau}}{\text{d}\boldsymbol{P}_{\tau}}=\frac{1}{\boldsymbol{P}_{\tau}\left(1-\boldsymbol{P}_{\tau}\right)}
\label{Equ14}
\end{equation}

Therefore, the simplified result of Equation~\ref{Equ12} is derived as:

\begin{equation}
\nabla_{\boldsymbol{P}_{\tau}}\boldsymbol{P}=\frac{1}{\boldsymbol{P}_{\tau}-1}\left(\boldsymbol{P}_1, \dots, \boldsymbol{P}_{y}, \dots, \boldsymbol{P}_{\tau}-1, \dots, \boldsymbol{P}_{K} \right)
\label{Equ15}
\end{equation}

To verify the relationship between the two optimization directions, we calculate the inner product and cosine similarity of the gradients of the example's probability distribution P along these two directions, which are given in Equation~\ref{Equ16} and Equation~\ref{Equ17}, respectively.

\begin{equation}
\nabla_{-\boldsymbol{P}_y}\boldsymbol{P}\cdot\nabla_{\boldsymbol{P}_{\tau}}\boldsymbol{P}=\frac{1}{\left(1-\boldsymbol{P}_y\right)\left(1-\boldsymbol{P}_{\tau}\right)}\left(\boldsymbol{P}_y+\boldsymbol{P}_\tau-\Phi\right).
\label{Equ16}
\end{equation}

\begin{equation}
\cos{\left(\nabla_{-\boldsymbol{P}_y}\boldsymbol{P},\nabla_{\boldsymbol{P}_{\tau}}\boldsymbol{P}\right)}=\frac{\boldsymbol{P}_y+\boldsymbol{P}_\tau-\Phi}{\sqrt{\left(\Phi-2\boldsymbol{P}_y+1\right)\left(\Phi-2\boldsymbol{P}_\tau+1\right)}}
\label{Equ17}
\end{equation}

In Equation~\ref{Equ16} and Equation~\ref{Equ17}, \(\Phi\) denotes the sum of the squares of the probabilities of all labels, which is expressed as follows:

\begin{equation}
\Phi=\sum_{k=1}^{K}{{\boldsymbol{P}_k}^2}.
\label{Equ18}
\end{equation}

It can be proven that the gradient inner product in Equation~\ref{Equ16} is always greater than 0, and the cosine similarity in Equation~\ref{Equ17} is always less than 1. Therefore, the two optimization directions in generation objective~\ref{Equ4} are neither mutually independent nor completely consistent; instead, they exhibit partial correlation.

\section{Projection Method Under the \(\bm{\ell}_2\)-Norm}
\label{SecB}

The gradient matrix is first normalized with \(\bm{\ell}_2\) normalization (i.e., Euclidean normalization). This operation scales the gradient matrix to a matrix with a unit norm, namely, making its \(\bm{\ell}_2\)-norm equal to 1 while preserving the original direction of the gradient, which is formulated as follows:

\begin{equation}
Grad=\frac{Grad}{{\|Grad\|}_2+\zeta}.
\label{Equ25}
\end{equation}

\begin{table*}[t]
\caption{Attack success rate of SDM with various hyperparameter combinations under 10 total iterations.}
\label{Tab6}
  \centering
  \newcolumntype{C}{>{\centering}X}
  \newcolumntype{L}[1]{>{\centering\arraybackslash}p{#1}}
  \scalebox{0.9}{
  \begin{tabularx}{0.95\linewidth}{C|*{2}{L{1.25cm}}|*{2}{L{1.25cm}}|*{2}{L{1.25cm}}|*{2}{L{1.25cm}}}
    \toprule
    \multirow{3}{*}{\makebox[0pt][c|]{Combination}} & \multicolumn{4}{c|}{CIFAR-10} & \multicolumn{4}{c}{CIFAR-100}\\
    & \multicolumn{2}{c|}{ResNet-34} & \multicolumn{2}{c|}{WideResNet-28-10} & \multicolumn{2}{c|}{ResNet-34} & \multicolumn{2}{c}{WideResNet-28-10}\\
    & \makebox[0pt][c]{VAT} & \makebox[0pt][c]{MART} & \makebox[0pt][c]{VAT} & \makebox[0pt][c]{MART} & \makebox[0pt][c]{VAT} & \makebox[0pt][c]{MART} & \makebox[0pt][c]{VAT} & \makebox[0pt][c]{MART}\\
    \midrule
    \makebox[0pt][c]{\ding{172}} & 55.10 & 48.22 & 56.38 & 51.39 & 76.41 & 72.58 & 77.08 & 73.34\\
    \makebox[0pt][c]{\ding{173}} & 56.87 & 52.78 & 58.37 & 56.62 & 79.17 & 76.78 & 79.89 & 77.25\\
    \bottomrule
  \end{tabularx}
  }
\end{table*}

\begin{table*}[t]
\caption{Attack success rate of SDM with various hyperparameter combinations under 20 total iterations.}
\label{Tab7}
  \centering
  \newcolumntype{C}{>{\centering}X}
  \newcolumntype{L}[1]{>{\centering\arraybackslash}p{#1}}
  \scalebox{0.9}{
  \begin{tabularx}{0.95\linewidth}{C|*{2}{L{1.25cm}}|*{2}{L{1.25cm}}|*{2}{L{1.25cm}}|*{2}{L{1.25cm}}}
    \toprule
    \multirow{3}{*}{\makebox[0pt][c|]{Combination}} & \multicolumn{4}{c|}{CIFAR-10} & \multicolumn{4}{c}{CIFAR-100}\\
    & \multicolumn{2}{c|}{ResNet-34} & \multicolumn{2}{c|}{WideResNet-28-10} & \multicolumn{2}{c|}{ResNet-34} & \multicolumn{2}{c}{WideResNet-28-10}\\
    & \makebox[0pt][c]{VAT} & \makebox[0pt][c]{MART} & \makebox[0pt][c]{VAT} & \makebox[0pt][c]{MART} & \makebox[0pt][c]{VAT} & \makebox[0pt][c]{MART} & \makebox[0pt][c]{VAT} & \makebox[0pt][c]{MART}\\
    \midrule
    \makebox[0pt][c]{\ding{172}} & 58.10 & 53.94 & 59.75 & 57.97 & 80.35 & 77.73 & 81.13 & 78.16\\
    \makebox[0pt][c]{\ding{173}} & 58.15 & 54.15 & 59.85 & 58.25 & 80.36 & 77.98 & 81.20 & 78.28\\
    \makebox[0pt][c]{\ding{174}} & 58.34 & 54.21 & 59.94 & 58.29 & 80.40 & 77.91 & 81.24 & 78.32\\
    \bottomrule
  \end{tabularx}
  }
    \vspace{3pt}
\end{table*}

\begin{table*}[t]
\caption{Attack success rate of SDM with various hyperparameter combinations under 50 total iterations.}
\label{Tab8}
  \centering
  \newcolumntype{C}{>{\centering}X}
  \newcolumntype{L}[1]{>{\centering\arraybackslash}p{#1}}
  \scalebox{0.9}{
  \begin{tabularx}{0.95\linewidth}{C|*{2}{L{1.25cm}}|*{2}{L{1.25cm}}|*{2}{L{1.25cm}}|*{2}{L{1.25cm}}}
    \toprule
    \multirow{3}{*}{\makebox[0pt][c|]{Combination}} & \multicolumn{4}{c|}{CIFAR-10} & \multicolumn{4}{c}{CIFAR-100}\\
    & \multicolumn{2}{c|}{ResNet-34} & \multicolumn{2}{c|}{WideResNet-28-10} & \multicolumn{2}{c|}{ResNet-34} & \multicolumn{2}{c}{WideResNet-28-10}\\
    & \makebox[0pt][c]{VAT} & \makebox[0pt][c]{MART} & \makebox[0pt][c]{VAT} & \makebox[0pt][c]{MART} & \makebox[0pt][c]{VAT} & \makebox[0pt][c]{MART} & \makebox[0pt][c]{VAT} & \makebox[0pt][c]{MART}\\
    \midrule
    \makebox[0pt][c]{\ding{172}} & 58.63 & 54.35 & 60.19 & 58.43 & 80.55 & 77.90 & 81.39 & 78.35\\
    \makebox[0pt][c]{\ding{173}} & 58.57 & 54.33 & 60.19 & 58.42 & 80.53 & 77.91 & 81.38 & 78.34\\
    \makebox[0pt][c]{\ding{174}} & 58.62 & 54.35 & 60.19 & 58.42 & 80.55 & 77.91 & 81.40 & 78.35\\
    \makebox[0pt][c]{\ding{175}} & 58.65 & 54.36 & 60.19 & 58.44 & 80.57 & 77.93 & 81.41 & 78.37\\
    \bottomrule
  \end{tabularx}
  }
\end{table*}

\begin{table*}[t]
\caption{Attack success rates of SDM against diverse architectures on the ImageNet-1K dataset.}
\label{Tab9}
  \centering
  \newcolumntype{C}{>{\centering}X}
  \newcolumntype{L}[1]{>{\centering\arraybackslash}p{#1}}
  \scalebox{0.9}{
  \begin{tabularx}{0.95\linewidth}{C|L{1.25cm}|L{1.25cm}|L{1.55cm}|L{2.25cm}|L{1.95cm}|L{3.2cm}}
    \toprule
    \makebox[0pt][c]{Attacks} & \makebox[0pt][c]{ViT-B} & \makebox[0pt][c]{Swin-L} & \makebox[0pt][c]{XCiT-L12} & \makebox[0pt][c]{CLIP-ViT-B/32} & \makebox[0pt][c]{ConvNeXt-L} & \makebox[0pt][c]{RaWideResNet-101-2}\\
    \midrule
    \makebox[0pt][c]{None} & \makebox[0pt][c]{31.62} & \makebox[0pt][c]{21.38} & \makebox[0pt][c]{26.24} & \makebox[0pt][c]{48.76} & \makebox[0pt][c]{21.98} & \makebox[0pt][c]{26.56}\\
    \makebox[0pt][c]{C\&W} & \makebox[0pt][c]{87.28} & \makebox[0pt][c]{66.00} & \makebox[0pt][c]{76.42} & \makebox[0pt][c]{98.57} & \makebox[0pt][c]{64.04} & \makebox[0pt][c]{72.21}\\
    \makebox[0pt][c]{PGD} & \makebox[0pt][c]{90.87} & \makebox[0pt][c]{67.32} & \makebox[0pt][c]{79.61} & \makebox[0pt][c]{97.55} & \makebox[0pt][c]{67.78} & \makebox[0pt][c]{75.03}\\
    \makebox[0pt][c]{APGD} & \makebox[0pt][c]{91.22} & \makebox[0pt][c]{68.05} & \makebox[0pt][c]{79.85} & \makebox[0pt][c]{98.32} & \makebox[0pt][c]{68.19} & \makebox[0pt][c]{75.31}\\
    \makebox[0pt][c]{AA} & \makebox[0pt][c]{\textbackslash} & \makebox[0pt][c]{70.52} & \makebox[0pt][c]{\textbackslash} & \makebox[0pt][c]{\textbackslash} & \makebox[0pt][c]{\textbackslash} & \makebox[0pt][c]{\textbackslash}\\
    \makebox[0pt][c]{SDM} & \makebox[0pt][c]{92.27} & \makebox[0pt][c]{70.31} & \makebox[0pt][c]{80.83} & \makebox[0pt][c]{99.25} & \makebox[0pt][c]{69.44} & \makebox[0pt][c]{76.29}\\
    \bottomrule
  \end{tabularx}
  }
\end{table*}

In Equation~\ref{Equ25}, \(\zeta\) is added to prevent division by zero. Subsequently, the adversarial example is updated along the direction of the unit gradient with a step size of \(\alpha\), and the current adversarial perturbation is generated as:

\vspace{-3pt}
\begin{equation}
\boldsymbol{\delta}_{t-1}=\boldsymbol{x}'_{t-1}+\alpha\cdot\text{sign}(Grad)-\boldsymbol{x}.
\label{Equ26}
\end{equation}

Then, the \(\bm{\ell}_2\)-norm of the current adversarial perturbation is calculated. If it exceeds the adversarial perturbation budget, the perturbation is scaled down to the range within the perturbation budget; if it does not exceed the budget, no processing is performed:

\vspace{-3pt}
\begin{equation}
\boldsymbol{\delta}_t=\boldsymbol{\delta}_{t-1}\cdot\min\left(\frac{\epsilon}{{\left\|\boldsymbol{\delta}_{t-1}\right\|}_2},1\right).
\label{Equ27}
\end{equation}

Finally, the projected adversarial perturbation is added to the natural example to generate the adversarial example under the \(\bm{\ell}_2\)-norm:

\vspace{-2pt}
\begin{equation}
\boldsymbol{x}'_t=\boldsymbol{x}+\boldsymbol{\delta}_t.
\label{Equ28}
\end{equation}

\section{Hyperparameter Selection Experiments}
\label{SecC}

For the hyperparameters used in SDM, the optimization cycle is the top-level parameter. In general, its value should be lower than the number of optimization stages and the number of iterative steps. Thus, for a total number of iterations of 10, two sets of hyperparameter combinations are configured in this section: \ding{172} \((1,2,5)\) and \ding{173} \((1,5,2)\) for \(C\), \(N\), and \(T\) respectively. Subsequently, attack evaluations are performed on the defense methods VAT and MART under the CIFAR-10 and CIFAR-100 datasets, with ResNet-34 and WideResNet-28-10 employed as the target models. The evaluation results are presented in Table~\ref{Tab6}, where the columns for VAT and MART represent the attack success rates of the attack method against the corresponding defense.

The results show that for a total number of iterations of 10, the average attack success rate of hyperparameter combination \ding{173} is 4.54\% higher than that of combination \ding{172}. Under the limited experimental conditions, this demonstrates that combination \ding{173} \((1,5,2)\) is the optimal hyperparameter set for a total iteration count of 10.

Next, based on the optimal combination \((1,5,2)\) for 10 total iterations, three sets of hyperparameter combinations are designed for a total iteration count of 20: \ding{172} \((2,5,2)\), \ding{173} \((1,10,2)\), and \ding{174} \((1,5,4)\) for \(C\), \(N\), and \(T\) respectively. The experimental setup for the evaluations remains consistent with the aforementioned experiments, and the results are shown in Table~\ref{Tab7}.

It can be observed that for a total number of iterations of 20, the average attack success rate of hyperparameter combination \ding{174} is 0.19\% higher than that of combination \ding{172} and 0.05\% higher than that of combination \ding{173}. This indicates that combination \ding{174} \((1,5,4)\) is the optimal set for a total iteration count of 20.

Subsequently, the optimal combination \((1,5,4)\) for 20 total iterations is extended to derive hyperparameter sets for a total iteration count of 50 from three perspectives: increasing the iterative steps per stage to yield \ding{172} \((1,5,10)\), increasing the number of optimization stages to yield \ding{173} \((1,10,5)\), and increasing the number of optimization cycles to yield \ding{174} \((2,6,4)\) and \ding{175} \((2,5,5)\). The experimental setup is consistent with the previous evaluations, and the results are presented in Table~\ref{Tab8}.

In Table~\ref{Tab8}, the average attack success rate of hyperparameter combination \ding{175} \((2,5,5)\) is 0.02\% higher than that of combination \ding{172}, 0.03\% higher than that of combination \ding{173}, and 0.02\% higher than that of combination \ding{174}. Under the constraint of 50 total iterations, combination \ding{175} exhibits the strongest attack performance and is therefore identified as the optimal hyperparameter set for this iteration count.

In the subsequent experiments for screening and determining hyperparameter combinations, the optimal sets for the total iteration counts of 100, 200, 500, and 1000 are identified in turn, and the detailed process is not repeated herein. All optimal hyperparameter combinations are collated uniformly in Table~\ref{Tab2}.

\vspace{-1.5pt}
\section{Cross-Architecture Generalization \& Scalability on Large‑Scale Datasets}
\label{SecD}

Evaluating the generalization ability of SDM across various architectures and its scalability on large-scale datasets is critical to validating its universal performance. Towards this goal, we conduct a set of controlled comparative experiments. Specifically, we evaluate representative models including ViT-B, Swin-L and XCiT-L12 (typical Transformer architectures), CLIP-ViT-B/32 (a contrastive learning-based multimodal model), ConvNeXt-L (a modern convolutional neural network), and RaWideResNet-101-2 (a classic convolutional neural network) on the ImageNet-1K (ILSVRC 2012) dataset.

To guarantee experimental fairness, we employ official third-party pre-trained weights rather than retraining all models from scratch. For CLIP-ViT-B/32, we adopt the non-robust clip-vit-base-patch32 weights loaded from the Transformers library. For other models, we utilize robust pre-trained weights retrieved from RobustBench \cite{CroceASDFCM021}, namely Mo2022When\_ViT-B, Xu2024MIMIR\_Swin-L, Debenedetti2022Light\_XCiT-L12, Liu2023Comprehensive\_ConvNeXt-L and Peng2023Robust. Afterwards, we perform adversarial attacks via PGD, C\&W, APGD, SDM and AA with 100 iterations under the \(\bm{\ell}_\infty\) perturbation bound of \(8/255\). The corresponding attack success rates (\%) are illustrated in Table~\ref{Tab9}.

\begin{table}[t!]
\caption{Attack success rates of SDM-based composite methods.}
\label{Tab10}
  \centering
  \newcolumntype{C}{>{\centering}X}
  \newcolumntype{L}[1]{>{\centering\arraybackslash}p{#1}}
  \scalebox{0.84}{
  \begin{tabularx}{1.16\linewidth}{C|*{4}{L{0.57cm}}|*{4}{L{0.57cm}}}
    \toprule
    \multirow{2}{*}{\makebox[0pt][c|]{Attack}} & \multicolumn{4}{c|}{CIFAR-10} & \multicolumn{4}{c}{CIFAR-100}\\
    & \makebox[0pt][c]{VAT} & \makebox[0pt][c]{\scalebox{0.9}[1]{MART}} & \makebox[0pt][c]{HAT} & \makebox[0pt][c]{\scalebox{0.9}[1]{LOAT}} & \makebox[0pt][c]{VAT} & \makebox[0pt][c]{\scalebox{0.9}[1]{MART}} & \makebox[0pt][c]{HAT} & \makebox[0pt][c]{\scalebox{0.9}[1]{LOAT}}\\
    \midrule
    \makebox[0pt][c]{APGD} & \makebox[0pt][c]{57.28} & \makebox[0pt][c]{49.37} & \makebox[0pt][c]{47.66} & \makebox[0pt][c]{45.96} & \makebox[0pt][c]{78.20} & \makebox[0pt][c]{73.20} & \makebox[0pt][c]{71.13} & \makebox[0pt][c]{69.56}\\
    \makebox[0pt][c]{SDM} & \makebox[0pt][c]{58.79} & \makebox[0pt][c]{54.43} & \makebox[0pt][c]{52.92} & \makebox[0pt][c]{51.04} & \makebox[0pt][c]{80.63} & \makebox[0pt][c]{77.89} & \makebox[0pt][c]{76.76} & \makebox[0pt][c]{74.17}\\
    \makebox[0pt][c]{\scalebox{0.92}[1]{APGD-AA}} & \makebox[0pt][c]{58.70} & \makebox[0pt][c]{55.10} & \makebox[0pt][c]{53.24} & \makebox[0pt][c]{51.39} & \makebox[0pt][c]{80.41} & \makebox[0pt][c]{78.22} & \makebox[0pt][c]{77.05} & \makebox[0pt][c]{74.48}\\
    \makebox[0pt][c]{\scalebox{0.92}[1]{SDM-AA}} & \makebox[0pt][c]{59.02} & \makebox[0pt][c]{55.35} & \makebox[0pt][c]{53.78} & \makebox[0pt][c]{51.80} & \makebox[0pt][c]{80.78} & \makebox[0pt][c]{78.69} & \makebox[0pt][c]{77.52} & \makebox[0pt][c]{74.93}\\
    \bottomrule
  \end{tabularx}
  }
\end{table}

Experimental results show that SDM outperforms C\&W, PGD and APGD by average attack success rate margins of 3.98\%, 1.71\% and 1.24\%, respectively. In terms of the ensemble attack AA, it suffers from extremely high computational cost, costing 111.92 hours on a single NVIDIA L40S GPU, which is far more than the 3.12 hours consumed by SDM. Hence, we only evaluate AA on the Swin-L model, and its attack efficacy is on par with SDM. Extensive experiments verify that SDM can be seamlessly deployed on Transformer-based architectures (distinct from traditional CNNs) and large-scale benchmarks like ImageNet-1K, while retaining prominent performance advantages over existing peer methods.

\section{Attack Performance of SDM-Based Composite Methods}
\label{SecE}

The integrated attack method AA is composed of three individual attack methods, namely APGD, FAB, and Square \cite{And20}. Despite its strong attack performance, it incurs extremely high computational costs. To further enhance the performance of AA, we replace APGD with SDM to form SDM-AA (the original AA is referred to as APGD-AA for distinction). Subsequently, we evaluate these two composite methods under the constraints of \(\boldsymbol{\bm{\ell}}_\infty=8/255\) and \(\boldsymbol{\bm{\ell}}_2=1.0\), respectively, with the results presented in Table~\ref{Tab10}. It can be observed that the average attack performance of SDM is only 0.25\% lower than that of the integrated method APGD-AA, while its computational cost is merely 15\%–20\% of the latter. Moreover, after upgrading with SDM, SDM-AA achieves an average attack performance that is 0.41\% higher than APGD-AA, providing a new avenue for the accurate evaluation of model robustness under sufficient computational resources.

\section{Evaluation of Anti-Interference Capability}
\label{SecF}

\begin{figure}[t]
  \centering
  \includegraphics[width=0.92\linewidth]{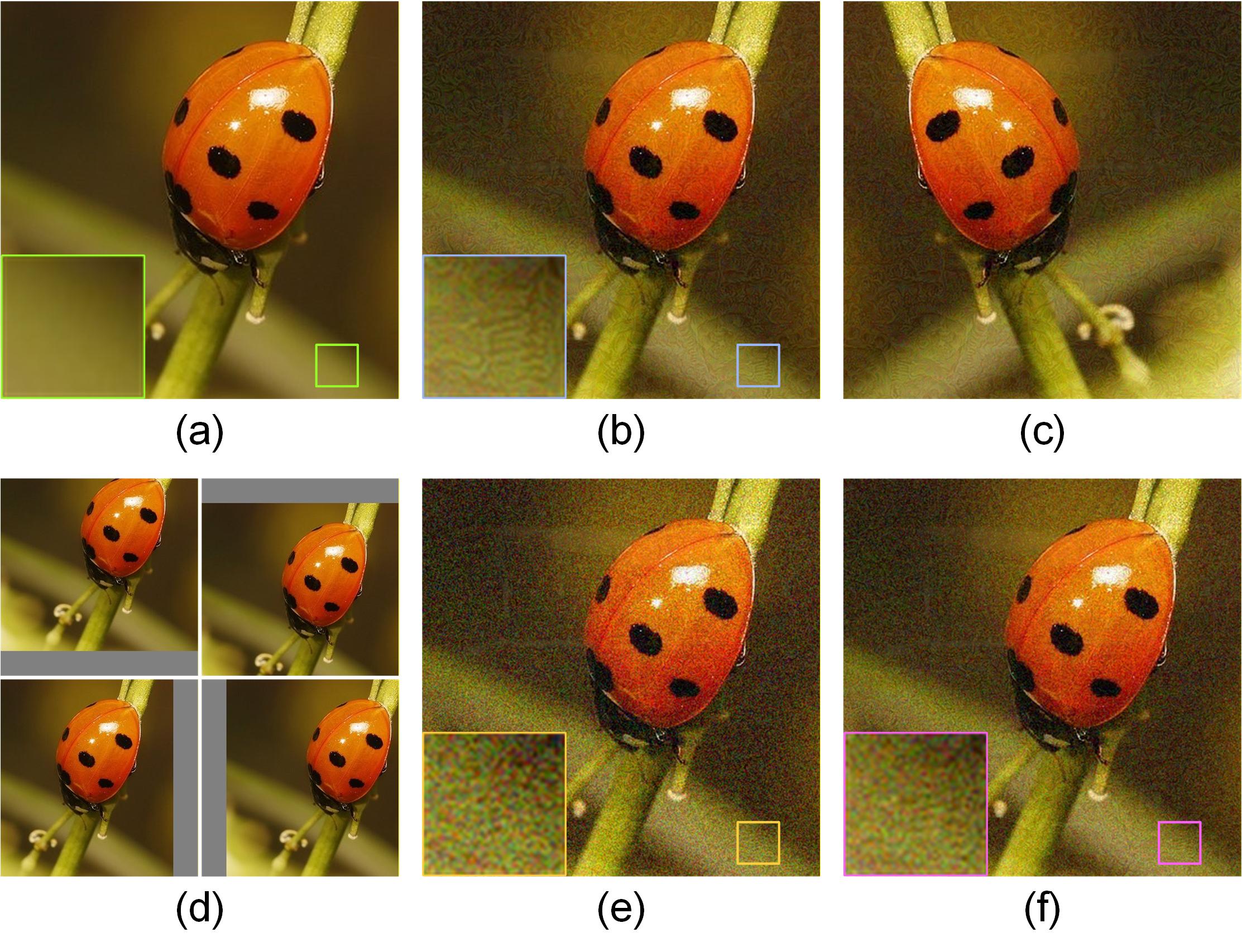}
  \caption{(a) A natural example; (b) an adversarial example; and adversarial examples perturbed by (c) horizontal flipping, (d) random translation, (e) uniform noise, and (f) Gaussian noise.}
  \label{Fig5}
\end{figure}

The anti-interference capability of adversarial examples refers to the ability to successfully launch adversarial attacks while resisting various external interferences, including noise, sensor distortion, active random interference from defense mechanisms, and computational errors. This capability is of great significance for the practical deployment of adversarial attack methods. To accurately evaluate this capability, corresponding attack experiments are designed in this section. 

As illustrated in Figure~\ref{Fig5}, for the generated raw adversarial examples (Figure~\ref{Fig5}(b), where the adversarial perturbation budget is greatly increased for better visualization), four types of interference effects are first designed in this section: horizontal mirror flipping (Figure~\ref{Fig5}(c)), random translation of the image by 12.5\% of its size (4 pixels on the CIFAR dataset) in one of the four directions (up, down, left, right) (Figure~\ref{Fig5}(d)), addition of uniform noise (UN) distributed over the interval \([-0.15,0.15]\) (with the pixel value range of adversarial examples being \([0,1]\)) (Figure~\ref{Fig5}(e)), and addition of Gaussian noise (GN) with a mean of 0 and a standard deviation of 0.05 (Figure~\ref{Fig5}(f)).

\begin{figure}[t]
  \centering
  \includegraphics[width=0.92\linewidth]{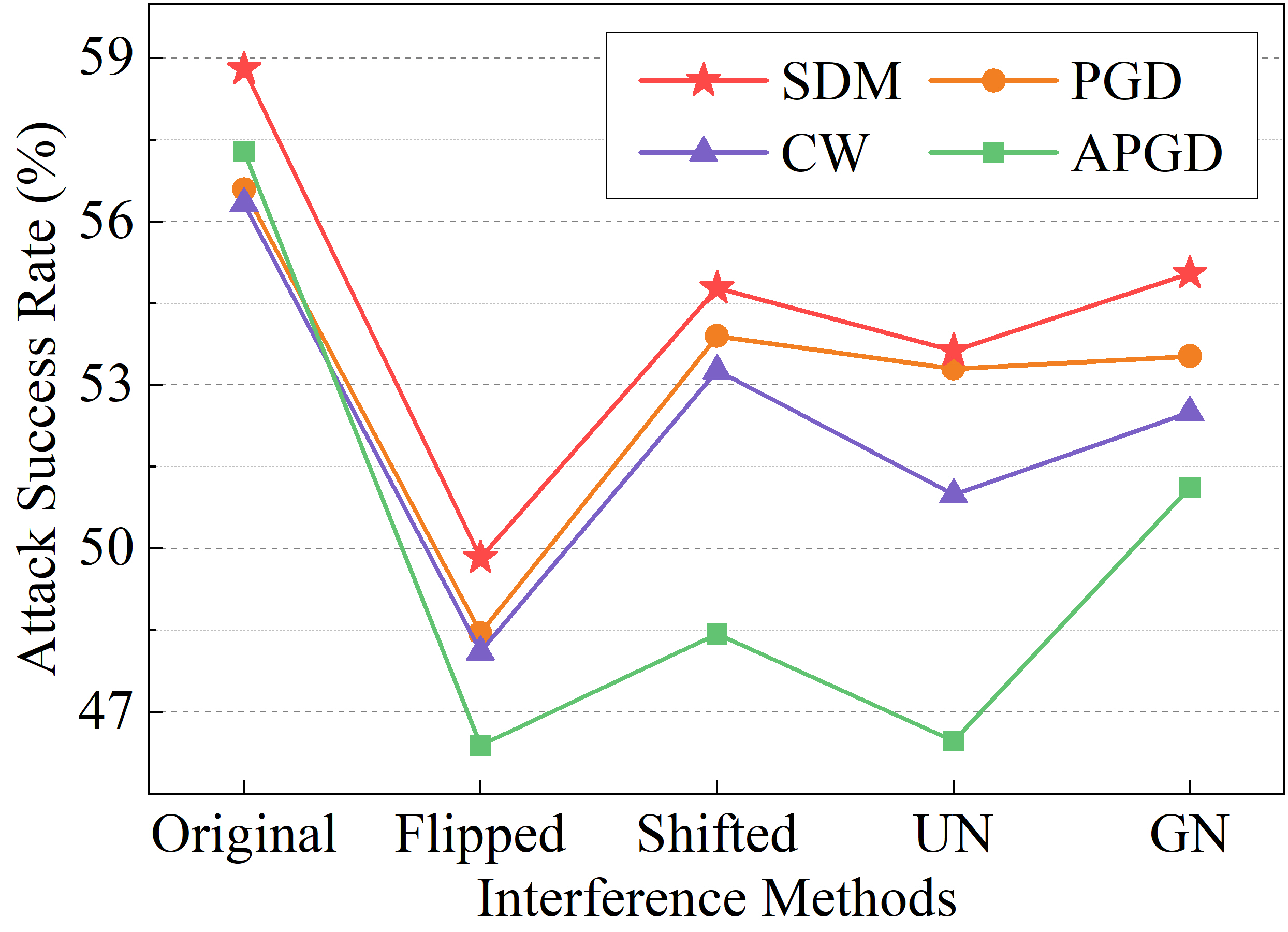}
  \caption{Attack success rates of various attack methods under different interferences with a \(\bm{\ell}_{\infty}\)-norm constraint.}
  \label{Fig6}
\end{figure}

Subsequently, on the CIFAR-10 dataset, adversarial attacks are launched against the defense method VAT using PGD, C\&W, APGD and the proposed SDM. Next, the four aforementioned interference effects are separately applied to the generated raw adversarial examples. Both the raw adversarial examples and the interference-added adversarial examples are input into the ResNet-34 target model, and their respective attack success rates are recorded after classification and recognition. The total number of iterations for all attack methods is set to 500. Under the \(\bm{\ell}_{\infty}\)-norm constraint, the adversarial perturbation budget and perturbation step size are \(8/255\) and \(2/255\), respectively; under the \(\bm{\ell}_2\)-norm constraint, the adversarial perturbation budget and perturbation step size are set to 1 and 0.2, respectively.

\begin{figure}[t]
  \centering
  \includegraphics[width=0.92\linewidth]{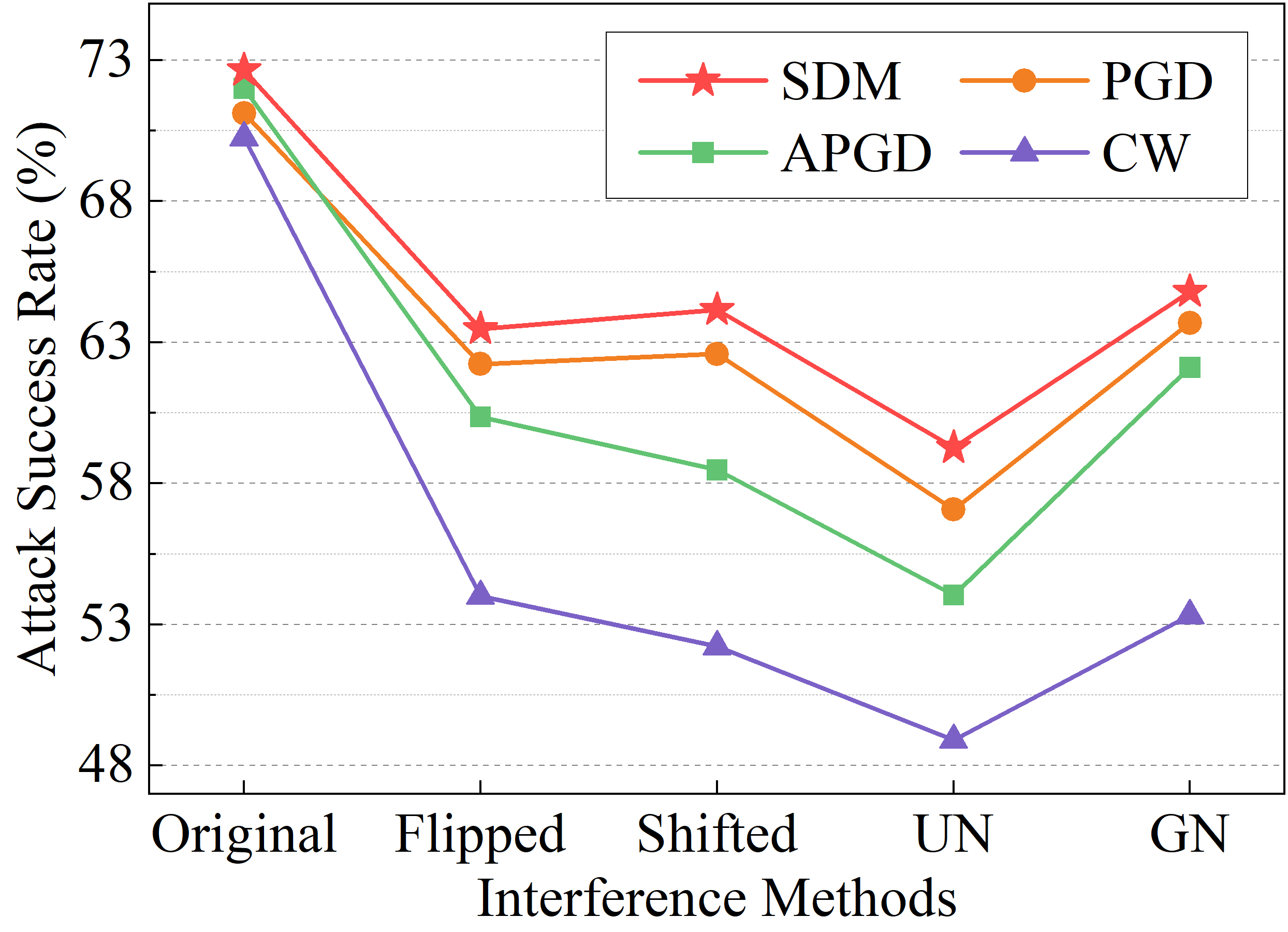}
  \caption{Attack success rates of various attack methods under different interferences with a \(\bm{\ell}_2\)-norm constraint.}
  \label{Fig7}
\end{figure}

The experimental results of the anti-interference capability under the \(\bm{\ell}_{\infty}\)-norm constraint are presented in Figure~\ref{Fig6}. It can be observed that under the \(\bm{\ell}_{\infty}\)-norm constraint, the adversarial examples generated by APGD exhibit the poorest anti-interference capability, with each type of interference causing a substantial drop in its attack success rate. This indicates that the adversarial perturbations generated by APGD under the \(\bm{\ell}_{\infty}\)-norm constraint have poor stability, and their perturbation structures are susceptible to damage from external interferences, thereby degrading adversarial performance. In contrast, the proposed SDM maintains the highest adversarial performance under all interference effects, which demonstrates that the adversarial perturbation structures it generates under the \(\bm{\ell}_{\infty}\)-norm constraint are stable and not easily damaged. The experimental results of the anti-interference capability under the \(\bm{\ell}_2\)-norm constraint are shown in Figure~\ref{Fig7}. It can be found that under the \(\bm{\ell}_2\)-norm constraint, the adversarial examples generated by the C\&W method have the worst anti-interference capability, with all types of interference leading to a significant decline in its adversarial performance. This reveals that the adversarial perturbations generated by the C\&W method under the \(\bm{\ell}_2\)-norm constraint are vulnerable to external interferences, which in turn impairs their adversarial performance. By contrast, the proposed SDM continues to achieve the highest anti-interference capability under the \(\bm{\ell}_2\)-norm constraint.

In summary, the adversarial examples generated by SDM under the constraints of the two involved norms (\(\bm{\ell}_{\infty}\) and \(\bm{\ell}_2\)) possess relatively stable adversarial structures and can maintain a high level of adversarial performance under multiple types of interferences. This result demonstrates the superior anti-interference capability of SDM, which translates to the strongest practical applicability in realistic environments.

\vspace{5pt}
\section{Analysis of Computational Complexity}
\label{SecG}

This section supplements Section~\ref{Sec57} and conducts a detailed computational complexity analysis of SDM. The classic PGD method mainly consists of data forward propagation, gradient backward propagation and CE loss computation. SDM adopts identical forward and backward propagation procedures as PGD; the primary distinction is that SDM uses the DPDR loss and incorporates probability ranking strategies.

In each iterative computation, let the batch size, input dimension \((C, H, W)\) and class number of the dataset be \(B\), \((C, H, W)\) and \(K\), respectively. The network depth and \scalebox{0.98}[1]{average channel quantity are denoted as \(L\) and \(D\). The com-}

\phantom{0}

\phantom{0}

\phantom{0}

\phantom{0}

\phantom{0}

\phantom{0}

\phantom{0}

\phantom{0}

putational complexity of PGD and SDM in each operation module is illustrated in Table~\ref{Tab11}.

It is observed that the DPDR loss introduces an extra ranking-related term \(O(B\cdot K\log K)\), which is nearly insignificant when compared to the major computational cost \(O(B\cdot C\cdot H\cdot W\cdot L\cdot D)\), as the dominant cost is 5 to 6 orders of magnitude larger. Hence, SDM possesses theoretically comparable computational complexity to PGD. To validate this finding, we perform experiments with 100 iterations under the \(\bm{\ell}_\infty\) constraint for PGD, APGD and SDM. We carry out evaluations on ResNet-34 trained on CIFAR-10 (32×32) and ViT-B trained on ImageNet-1K, with batch sizes (BS) set to 32 and 128 separately. All experiments are run on an NVIDIA L40S GPU and repeated five times. We record the step-wise runtime in milliseconds, and the statistical results are presented in Table~\ref{Tab12}.  

\begin{table}[t!]
\caption{Computational complexity of PGD and SDM.}
\label{Tab11}
  \centering
  \newcolumntype{C}{>{\centering}X}
  \newcolumntype{L}[1]{>{\centering\arraybackslash}p{#1}}
  \scalebox{0.83}{
  \begin{tabularx}{1.18\linewidth}{C|L{3.28cm}|L{3.82cm}}
    \toprule
    \makebox[0pt][c]{Item} & \makebox[0pt][c]{PGD} & \makebox[0pt][c]{SDM}\\
    \midrule
    \makebox[0pt][c]{Forward} & \makebox[0pt][c]{\(O(B\cdot C\cdot H\cdot W\cdot L\cdot D)\)} & \makebox[0pt][c]{\(O(B\cdot C\cdot H\cdot W\cdot L\cdot D)\)}\\
    \makebox[0pt][c]{Backward} & \makebox[0pt][c]{\(O(B\cdot C\cdot H\cdot W\cdot L\cdot D)\)} & \makebox[0pt][c]{\(O(B\cdot C\cdot H\cdot W\cdot L\cdot D)\)}\\
    \makebox[0pt][c]{CE Loss} & \makebox[0pt][c]{\(O(B\cdot K)\)} & \makebox[0pt][c]{\textbackslash}\\
    \makebox[0pt][c]{DPDR Loss} & \makebox[0pt][c]{\textbackslash} & \makebox[0pt][c]{\(O(B\cdot K)+O(B\cdot K\log K)\)}\\
    \bottomrule
  \end{tabularx}
  }
\end{table}

\begin{table}[t!]
\caption{Practical runtime (ms) of various attack methods.}
\label{Tab12}
  \centering
  \newcolumntype{C}{>{\centering}X}
  \newcolumntype{L}[1]{>{\centering\arraybackslash}p{#1}}
  \scalebox{0.84}{
  \begin{tabularx}{1.16\linewidth}{C|L{1.53cm}L{1.66cm}|L{1.66cm}L{1.8cm}}
    \toprule
    \multirow{2}{*}{\makebox[0pt][c|]{Attack}} & \multicolumn{2}{c|}{CIFAR-10} & \multicolumn{2}{c}{ImageNet-1K}\\
    & \makebox[0pt][c]{BS = 32} & \makebox[0pt][c]{BS = 128} & \makebox[0pt][c]{BS = 32} & \makebox[0pt][c]{BS = 128}\\
    \midrule
    \makebox[0pt][c]{PGD} & \makebox[0pt][c]{7.30 \(\pm\) 0.21} & \makebox[0pt][c]{12.24 \(\pm\) 0.13} & \makebox[0pt][c]{92.21 \(\pm\) 5.23} & \makebox[0pt][c]{376.79 \(\pm\) 9.86}\\
    \makebox[0pt][c]{APGD} & \makebox[0pt][c]{8.36 \(\pm\) 0.31} & \makebox[0pt][c]{11.69 \(\pm\) 0.16} & \makebox[0pt][c]{80.10 \(\pm\) 6.72} & \makebox[0pt][c]{356.02 \(\pm\) 12.34}\\
    \makebox[0pt][c]{SDM} & \makebox[0pt][c]{8.15 \(\pm\) 0.26} & \makebox[0pt][c]{13.33 \(\pm\) 0.14} & \makebox[0pt][c]{95.02 \(\pm\) 5.18} & \makebox[0pt][c]{385.50 \(\pm\) 11.75}\\
    \bottomrule
  \end{tabularx}
  }
  \vspace{-13.8pt}
\end{table}

On the CIFAR-10 dataset, the average runtime of SDM is 9.92\% and 7.13\% higher than that of PGD and APGD respectively. On ImageNet-1K, the increased ratios are 2.46\% and 8.28\%. Although equipped with a more elaborate optimization paradigm, SDM only brings slight additional overhead within the range of 5\% to 9\%. In addition, we think that such extra consumption is not derived from the core logic of the SDM algorithm, but mainly results from frequent tensor initialization and index operations in DPDR. In future research, we will focus on minimizing the number of memory I/O operations to further accelerate the algorithm.

\end{document}